\documentclass[dvipsnames]{article}
\usepackage{colm2024_conference}

\usepackage{booktabs}
\usepackage{enumitem}
\usepackage{wrapfig}
\usepackage{graphicx}
\usepackage{caption}
\usepackage{subcaption}
\usepackage{multirow}
\usepackage{tabularx}
\usepackage{makecell}
\usepackage{adjustbox}
\usepackage{colortbl}
\usepackage{rotating}
\usepackage{xltabular}

\usepackage[utf8]{inputenc}
\usepackage[T1]{fontenc}
\usepackage{microtype}
\usepackage{setspace}
\usepackage[normalem]{ulem}
\usepackage{xspace}
\usepackage{url}
\usepackage{xurl}

\usepackage{mathtools}
\usepackage{siunitx}
\usepackage{nicefrac}


\usepackage{algorithm}
\usepackage{algpseudocode}

\usepackage{pgfplots}
\pgfplotsset{compat=1.18}
\usepackage{tikz}
\usetikzlibrary{er,positioning,bayesnet,tikzmark}

\usepackage{listings}
\usepackage[raster,skins,most]{tcolorbox}
\usepackage{fontawesome}
\usepackage[misc]{ifsym}
\usepackage{tipa}
\usepackage{fancyvrb}
\usepackage{color}

\definecolor{lightgray}{rgb}{0.9,0.9,0.9}
\definecolor{babygreen}{rgb}{0.85, 0.97, 0.85}
\definecolor{deepblue}{RGB}{48, 58, 82}
\definecolor{cadmiumgreen}{rgb}{0.0, 0.42, 0.24}
\definecolor{deepPurple}{HTML}{330066}
\definecolor{uclablue}{rgb}{0.15, 0.45, 0.68}

\newcommand\inb[1]{\colorbox{gray!20}{\lstinline|#1|}}

\usepackage{amsmath,amsfonts,bm}

\def\eqref#1{equation~\ref{#1}}

\def\1{\bm{1}}

\DeclareMathAlphabet{\mathsfit}{\encodingdefault}{\sfdefault}{m}{sl}
\SetMathAlphabet{\mathsfit}{bold}{\encodingdefault}{\sfdefault}{bx}{n}

\usepackage{pifont}
\newcommand{\cmark}{\ding{51}}
\newcommand{\xmark}{\ding{55}}

\hypersetup{
    breaklinks,
    citecolor=uclablue,
    colorlinks=true,
}

\newcommand{\benchname}{EIBench\xspace}
\newcommand{\methodname}{CTC-GRPO\xspace}

\newcommand{\scCharm}{\textsc{Charm}\xspace}
\newcommand{\scDefense}{\textsc{Defense}\xspace}
\newcommand{\scRepair}{\textsc{Repair}\xspace}
\newcommand{\scSupport}{\textsc{Support}\xspace}

\title{%
EIBench: A Simulator-Based Benchmark and Turn-Credit RL \\
for Emotion Management
}

\newcommand{\symboletongyi}{\,\raisebox{-0.4ex}{\includegraphics[height=2ex]{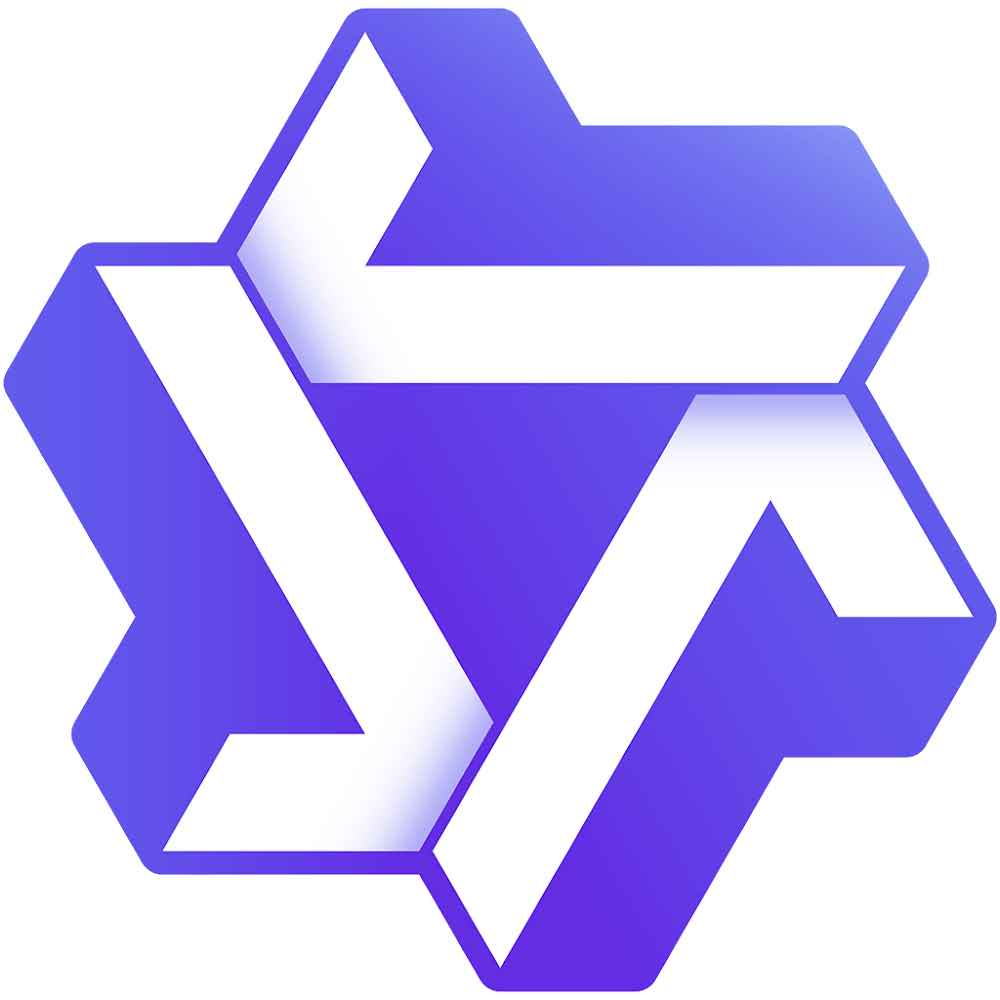}}}

\author{%
Rongzhi Zhu$^{1}$ \quad
Xiang Huang$^{2}$ \quad
Yuchuan Wu$^{2}$ \quad
Rui Wang$^{2}$ \quad
Zequn Sun$^{1,\dagger}$ \\
Tao Ren$^{2}$ \quad
Weiyao Luo$^{2}$ \quad
Bingxue Qiu$^{2}$ \quad
Jieping Ye$^{2}$ \quad
Yongbin Li$^{2,\dagger}$ \quad
Wei Hu$^{1}$ \\[6pt]
\normalsize
$^{1}$State Key Laboratory for Novel Software Technology, Nanjing University, \\
Nanjing 210023, China \\
$^{2}$Qwen-Character Team\symboletongyi, Alibaba Group \\[3pt]
{\small $^{\dagger}$Corresponding authors.}
}

\begin{document}
\maketitle

\begin{abstract}
Emotional intelligence (EI) in Large Language Models (LLMs) is often evaluated through static understanding tasks or single-response dialogue generation.
However, emotion management is interactive: a good model should not only recognize a user's emotion, but also improve the user's emotional and relational state over several turns.
We introduce \benchname{}, a simulator-based benchmark for interactive emotion management.
\benchname{} contains 2{,}222 scenarios, with 2{,}009 for training and 213 for held-out testing.
The scenarios are organized by a 2$\times$2 taxonomy covering Support, Defense, Repair, and Charm, which together capture different forms of support, boundary maintenance, trust repair, and rapport building.
In each scenario, an LLM simulator plays the user, updates an emotion--relation state after each turn, and maps the final state to an anchor-based score.
This design makes \benchname{} both an evaluation benchmark and a training environment: the final state gives the outcome reward, while the per-turn state updates provide dense feedback for RL.
We evaluate 15 open- and closed-source LLMs.
Current models perform well on support and rapport-building scenes, but struggle with boundary maintenance under user pressure.
To improve the EI ability of LLMs, we propose Centered Turn-Credit GRPO (\methodname{}), a GRPO extension that reuses the simulator's per-turn state updates as dense turn-level feedback while preserving the final outcome reward.
\methodname{} improves Qwen3-8B from $-22.4$ to $+22.4$ on \benchname{} and also improves on out-of-distribution evaluations including SAGE~(+12.4) and EQBench3~(+20.9\%).
Our results show that simulator-tracked user states can support both evaluation and training for multi-turn emotion management.
\end{abstract}

\section{Introduction}
\label{sec:intro}

Large language models (LLMs) have advanced rapidly with recent scaling and post-training breakthroughs \citep{gpt4,gemini,deepseek,qwen3}. These gains are exceptionally prominent in complex reasoning domains such as mathematics \citep{deepseekmath,mathimo} and coding \citep{qwen3coder,deepseekcoder}. 
A key driver of these gains is reinforcement learning with verifiable rewards (RLVR), where correctness labels provide clear, automated feedback and make reward optimization relatively straightforward \citep{rlvr,simplerl,tongyideepresearch}.


However, for subjective tasks such as social and emotional interaction, there is often no single correct response. 
What matters is how the model's actions shape and shift the other person's emotional state across multiple 
turns. 
Mayer and Salovey \citep{ei} conceptualize this ability as emotional intelligence (EI) and distinguish four hierarchical layers: \emph{perceiving}  emotions, \emph{using} them to facilitate thought, \emph{understanding} their structure, and \emph{managing} them in oneself and others. 
The first three  abilities can often be assessed from a fixed context---identifying which emotion  is present, explaining why it arises, or examining how it informs reasoning.  
Emotion management, however, is fundamentally different: it is a state-changing problem in which a model's responses are judged by how  effectively they guide the other person's emotional and relational state over time.

We observe that most EI evaluations still focus on emotion understanding or narrow forms of emotional interaction, rather than broad emotion management across multi-turn dialogue.
Early datasets such as Empathetic Dialogues \citep{towards_empathetic} and ESConv \citep{esconv} treat emotional dialogue as supervised learning over annotated conversations.
Recent benchmarks move toward more open-ended evaluation: EQBench3 \citep{eqbench} asks models to infer mental states and choose appropriate responses, while SOTOPIA \citep{sotopia} evaluates goal completion in multi-agent social scenarios.
SAGE \citep{sage} is closer to our setting, as an LLM simulator plays a support seeker and the tested model provides support over a full dialogue.
However, these settings still do not fully cover emotion management.
SOTOPIA focuses on goal completion rather than improving the other person's emotional and relational state, while SAGE focuses mainly on support-seeking dialogues.
This leaves common cases such as handling pressure, repairing trust after model-side mistakes, and proactively building a relationship less studied.

To address this gap, we build an interactive benchmark for emotion
management. In our setup, two LLMs are engaged in a multi-turn dialogue, and the
score is based on how the simulated user's state changes over time. We track both emotion and relation because improving the user's mood and building a better relationship are related but not identical goals. The
same state trajectory also provides turn-level feedback for RL training.
To cover different forms of emotion management, we organize the scenarios
with a 2$\times$2 taxonomy grounded in the EI literature. The taxonomy separates the target of regulation from the source of the situation,
yielding four scene types: \scSupport{}, \scDefense{}, \scRepair{}, and \scCharm{}.

Overall, we make three main contributions:
\begin{itemize}[leftmargin=1.2em,itemsep=2pt]

\item \textbf{EIBench}: We introduce a multi-turn benchmark for emotion
management, covering \scSupport{}, \scDefense{}, \scRepair{}, and
\scCharm{} under a 2$\times$2 taxonomy based on the target of regulation
and the source of the situation.

\item \textbf{Leaderboard and Validation}: We evaluate 15 open- and 
closed-source LLMs on EIBench. 
We report rankings under two independent simulators and validate them with a human pairwise arena.

\item \textbf{Centered Turn-Credit GRPO (\methodname{}):}
We propose \methodname{} for multi-turn dialogue optimization.
It uses the simulator's per-turn states as process rewards with a centered credit term to redistribute turn‑level credit without altering the outcome signal.
The trained models also improve on out-of-distribution evaluations including
EQBench3 and SAGE.
\end{itemize}

\section{Related Work}
\label{sec:related}

\subsection{EI Benchmarks}
\label{sec:related:bench}

Early NLP work on EI used static datasets such as Empathetic Dialogues \citep{towards_empathetic}, ESConv \citep{esconv}, and PsyQA \citep{psyqa} to teach models supportive replies under a fixed context.
As LLMs strengthened, benchmarks such as EQBench3 \citep{eqbench}, EmoBench \citep{emobench}, ToMBench \citep{tombench}, and SI-Bench \citep{sibench} moved into richer scenarios for emotional understanding and social reasoning.
Related role-play benchmarks such as CoSER \citep{coser} further evaluate character-grounded dialogue generation.
However, they are still offline or end-to-end: the model cannot affect what the user says next.

Recent benchmarks add dynamic interaction. AgentSense \citep{agentsense} and SOTOPIA \citep{sotopia} place multiple agents into role-bound goals.
SAGE \citep{sage} and Echo-N1 \citep{echo} use an LLM simulator that plays an evolving support seeker.
They are dynamic but narrow---goal completion or support-only. 
Our benchmark spans all four scene types (\scSupport, \scDefense, \scRepair, \scCharm) and shares the same simulator between evaluation and RL training.
This difference is important because support-only evaluation does not test whether a model can handle pressure, recover from its own mistakes, or build rapport when no explicit distress is given.

\begin{table*}[t]
\centering
\small
\setlength{\tabcolsep}{5pt}
\begin{tabular}{@{}lcccccc@{}}
\toprule
Dataset & Size & Multi-turn & Sim. & State Score & Scope & Score Dim. \\
\midrule
ESConv \cite{esconv}      & 195 & \xmark & \xmark & \xmark & Emotion support    & -- \\
EQBench3 \cite{eqbench}   & 45  & \xmark & \xmark & \xmark & EQ reasoning         & EQ rubric \\
CoSER \cite{coser}        & 200 & \cmark & \xmark & \xmark & Role-play            & Role consistency \\
SOTOPIA \cite{sotopia}    & 90  & \cmark & \cmark & \xmark & Social interaction   & Goal \\
SAGE \cite{sage}          & 100 & \cmark & \cmark & \cmark & Emotion Support     & Emotion \\
\midrule
\benchname{}              & 213 & \cmark & \cmark & \cmark & Emotion management   & Emotion + Relation \\
\bottomrule
\end{tabular}
\caption{Benchmark comparison. Size is test-set size; Sim. is simulator interaction.}
\label{tab:benchmark_comparison}
\end{table*}

\subsection{Multi-Turn RL and Credit Assignment}
\label{sec:related:rl}

On-policy RL has become a central part of recent LLM post-training, especially after reasoning models such as DeepSeek-R1 \citep{deepseek}.
In the standard RLVR setting \citep{dapo,vapo,gspo}, a model receives a reward from the final answer, which works well for tasks with clear correctness criteria such as math and coding.
Recent extensions to agentic reasoning and tool use \citep{searchr1,gigpo} allow models to act over multiple steps, but the feedback is still often delayed or task-specific.
For subjective tasks without a fast verifier, recent work often uses LLM-as-a-judge scoring \citep{llmjudge,rlver,mapo} or pairwise ranking \citep{arenarl}.
These methods make open-ended optimization possible, but they usually provide only a terminal score.
This creates a credit assignment problem: a single final reward cannot tell which turn helped or hurt the interaction.

Recent methods densify feedback with reward models, turn-level judges, or task-specific process rewards \citep{grm,mapo,gigpo,autorefine}.
Our setting provides a simpler signal: the simulator already updates the user's emotion and relation state after each model reply.
We reuse these state changes as turn-level process rewards, without adding another judge model.
This gives dense feedback while keeping the final anchor-based outcome score unchanged for evaluation.

\section{\benchname for Emotion Management}
\label{sec:bench}

\begin{figure*}[t]
\centering
\includegraphics[width=\linewidth]{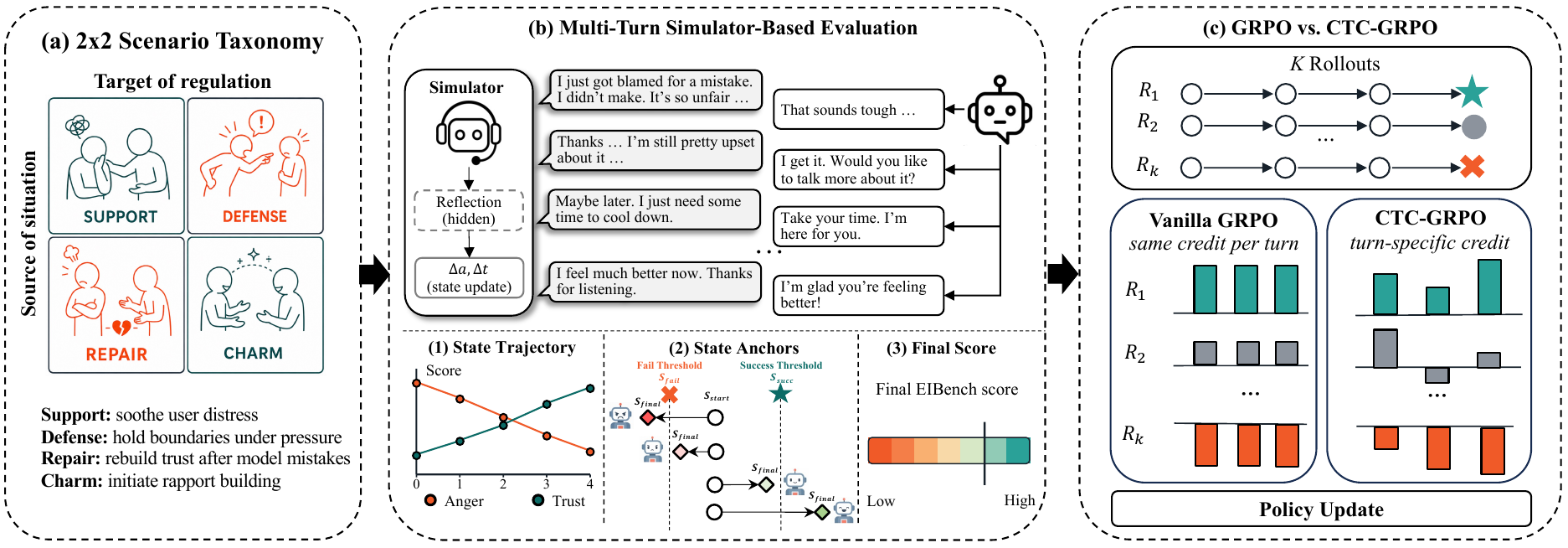}
\caption{Overview of \benchname{} and \methodname{}. \benchname{} organizes emotion-management scenarios with a 2$\times$2 taxonomy, evaluates models through multi-turn simulator interaction and anchor-based state scoring, and provides turn-level state changes for \methodname{} training.}
\label{fig:framework}
\end{figure*}

This section presents the design of \benchname{}.
Table~\ref{tab:benchmark_comparison} first compares \benchname{} with related benchmarks, and Figure~\ref{fig:framework} gives the overall workflow.
We then introduce the 2$\times$2 scenario taxonomy, describe scenario construction and state anchors, and define the multi-turn simulator-based scoring method.

\subsection{Foundations: A 2$\times$2 Scenario Taxonomy}
\label{sec:bench:taxonomy}

Our taxonomy is grounded in two ideas from emotion research.
Mayer and Salovey \citep{ei} define emotion management as both interpersonal and intrapersonal.
We use this distinction as the first axis: the target of regulation.
A scene is other-directed when the model mainly needs to improve the user's emotional or relational state, and self-directed when the main challenge is to regulate the model's own stance, boundary, or initiative.
This axis is not enough by itself.
For example, comforting a user after an external loss and repairing trust after a model mistake both aim to improve the user's state, but they require different actions.
We therefore add a second axis: the source of the situation, following appraisal theory \citep{emotion}.
 A scene is user-side when the need comes from the user or an external event. It is model-side when it comes from the model's own mistake or initiative.

Crossing the two axes gives the four scene types summarized in Table~\ref{tab:taxonomy}.
\scSupport{} covers externally caused user distress.
\scDefense{} covers user pressure, where the model must stay calm and maintain its boundary.
\scRepair{} covers model-side mistakes and trust rebuilding.
\scCharm{} covers model-initiated rapport building.
The two axes define the main interaction goal of each scene.
Other factors, such as emotional intensity, relationship history, power relation, and setting, mainly affect the difficulty or surface form.

\subsection{Dataset Construction}
\label{sec:bench:dataset}

\benchname{} contains 2{,}222 scenarios: 2{,}009 for training and 213 for held-out testing.
Each scenario is a multi-turn role-play between the model under test and an LLM simulator.

\subsubsection{Scenario Components}
\label{sec:bench:components}

A scenario has two role profiles and three state anchors on
emotion--relation axes. The \texttt{simulated-} \texttt{\_user\_profile} is given to the simulator and specifies the user's role, background, speaking style, rules, and goal. 
The \texttt{model\_profile} is given to the model under
test and specifies its role, known facts, and core goal.
Most scenarios start with an \texttt{opening\_line} from the simulated
user. 
The only exception is \scCharm{}, where the model under test speaks
first as the interaction is model-initiated. The three anchors define the start, successful, and failed states for scoring:
$s_{\mathrm{start}}$, $s_{\mathrm{succ}}$, and $s_{\mathrm{fail}}$.

\subsubsection{Scenario Generation and Anchor Setup}
\label{sec:bench:generation}
We first construct the test set from human-written and human-collected
scenario seeds. Each seed specifies the background story, speaker
relationship, trigger event, and desired outcome. Gemini-3.1-Pro is used
only to convert these seeds into our profile format; the core interaction
remains human written or human collected. We then manually check each
scenario for plausibility, profile consistency, and anchor quality.

\paragraph{Training scenario generation.}
After the test set is fixed, we build seed pools for training-set
generation. 
The pools include theme directions, scenario keywords, and three modifier dimensions: relationship, emotional intensity, and personality influence.
Appendix~\ref{app:bench-stats} gives the full seed-pool statistics.
The entries are written and expanded with manual editing and LLM assistance, and
vague or overlapping entries are removed. To generate a training
scenario, we sample one scenario keyword and one option from each
modifier dimension, then ask Gemini-3.1-Pro to expand the sampled
combination into a full scenario in our profile format.

To reduce train-test overlap, we use two safeguards. First, keyword
combinations that appear in the test set are removed from the training
pool. Second, after each training scenario is generated, we retrieve the
three most similar test scenarios with an embedding model and ask
Gemini-3.1-Pro to check whether it duplicates any of them. Duplicated
scenarios are rewritten.

\begin{table}[t]
\centering
\small
\setlength{\tabcolsep}{5pt}
\begin{tabular}{@{}llll@{}}
\toprule
Scene & Regulation target & Situation source & Main goal \\
\midrule
\scSupport{} & user emotion & user event & comfort and stabilize \\
\scDefense{} & model boundary and tone & user pressure & remain firm but warm \\
\scRepair{} & user trust & model mistake & acknowledge and rebuild trust \\
\scCharm{} & model rapport-building & model-initiated interaction & build rapport \\
\bottomrule
\end{tabular}
\caption{The four scene types in EIBench. The two middle columns correspond to the two axes of our taxonomy.}
\label{tab:taxonomy}
\end{table}

\paragraph{State anchors.}

Each scenario has three state anchors on the two scoring axes:
negative emotion intensity $a$ and relational state $t$.
The anchors are the start state $s_{\mathrm{start}}$,
the successful target state $s_{\mathrm{succ}}$,
and the failed state $s_{\mathrm{fail}}$.
They define where the dialogue starts, what counts as a realistic successful outcome,
and what counts as a clear failure.
These anchors make scores comparable across different scenes:
a final state is judged by its progress from the start state toward the successful or failed state, rather than by an absolute emotion or relation value.

To make the anchors more consistent, we use a behavior-based rubric with five score ranges for each axis.
The rubric also explains how the axes should be interpreted in different scenes:
negative emotion mainly refers to anger in \scDefense{} and \scRepair{},
sadness or despair in \scSupport{}, and aversion or defensiveness in \scCharm{}.
The full rubric and anchor-labeling system prompt are provided in Appendix~\ref{app:prompt}.
Gemini-3.1-Pro proposes the anchors under this rubric.
We use scene-level target ranges to keep $s_{\mathrm{succ}}$ reachable rather than idealized:
a successful state should be a plausible good outcome for that scene, not a perfect emotional endpoint. This is especially important for scenes where the conflict cannot be fully removed within a short dialogue.
In \scDefense{}, for example, a good outcome is not that the user becomes fully satisfied with the refusal.
Rather, the model should keep a necessary boundary while reducing the user's frustration and keeping the interaction open.
Similarly, in \scRepair{}, the user may not fully trust the model again after one conversation, but a successful dialogue should acknowledge the mistake, reduce anger, and restore enough trust for continued engagement.
Thus, the successful anchor represents a realistic improvement target under the scenario constraints, while the failed anchor represents a clear breakdown of the interaction.

We then use a second LLM verifier to check scenario consistency, target reachability, and anchor ordering.
It adjusts anchors only when they conflict with the scenario, make the target unreachable, or violate the required ordering.
Finally, we round all anchors to multiples of five and apply deterministic rules to keep them within $[0,100]$ and preserve the required gaps.

\paragraph{Statistics.}
Table~\ref{tab:bench:stats} reports per-scene train/test counts and anchor statistics.
The four scene types differ in both start and successful states.
\scCharm{} starts with low negative emotion and low relational state because the speakers have no prior relationship.
\scDefense{} and \scRepair{} start with high negative emotion and low relational state, while \scSupport{} starts with high negative emotion but a higher relational state.
Their successful states also differ: \scCharm{} aims to build a positive relation, \scDefense{} and \scRepair{} reduce negative emotion and improve relation under realistic limits, and \scSupport{} aims for the strongest relational endpoint.
By contrast, the failed states are consistently high in negative emotion and low in relational state, representing clear breakdowns of the interaction.
These differences show that the scenes capture distinct interaction goals rather than simple variants of the same task.

\begin{table}[t]
\centering
\small
\setlength{\tabcolsep}{3.2pt}
\renewcommand{\arraystretch}{0.92}
\begin{tabular}{@{}l c cc cc cc@{}}
\toprule
Scene & Train/Test
& \multicolumn{2}{c}{Start}
& \multicolumn{2}{c}{Success}
& \multicolumn{2}{c}{Fail} \\
\cmidrule(lr){3-4}\cmidrule(lr){5-6}\cmidrule(l){7-8}
& & $a$ & $t$ & $a$ & $t$ & $a$ & $t$ \\
\midrule
\scCharm{}   & 504/46 & $34{\pm}13$ & $17{\pm}8$  & $6{\pm}4$  & $57{\pm}6$  & $75{\pm}10$ & $1{\pm}3$ \\
\scDefense{} & 496/64 & $75{\pm}14$ & $16{\pm}11$ & $30{\pm}7$ & $46{\pm}7$  & $97{\pm}6$  & $0{\pm}2$ \\
\scRepair{}  & 503/52 & $77{\pm}10$ & $10{\pm}5$  & $28{\pm}6$ & $50{\pm}10$ & $97{\pm}6$  & $0{\pm}1$ \\
\scSupport{} & 506/51 & $75{\pm}9$  & $46{\pm}14$ & $33{\pm}8$ & $82{\pm}7$  & $97{\pm}6$  & $9{\pm}7$ \\
\bottomrule
\end{tabular}
\caption{Per-scene statistics for \benchname{}.
Anchors are mean$\pm$std on the training split; $a$ and $t$ denote negative emotion intensity and relational state.}
\label{tab:bench:stats}
\vspace{-0.5em}
\end{table}

\subsection{Simulator-Based Evaluation}
\label{sec:bench:simulator}

EIBench evaluates a model by simulating a multi-turn dialogue and tracking
two user states: negative emotion intensity $a_\tau\in[0,100]$ and
relational state $t_\tau\in[0,100]$. Lower $a_\tau$ and higher $t_\tau$
indicate better outcomes. The emotion axis measures how upset or
resistant the user is, while the relation axis measures whether the user
is willing to engage with or trust the model. We use two axes because
emotion relief and relationship improvement are not the same. The
simulator follows the \texttt{simulated\_user\_profile}, plays the user,
and after each model reply produces the next user reply, a state update
$\delta_\tau=(\Delta a_\tau,\Delta t_\tau)$, and a stop decision. The state starts from $s_{\mathrm{start}}=(a_{\mathrm{start}},t_{\mathrm{start}})$ and evolves as
$s_\tau=s_{\tau-1}+\delta_\tau$, clipped to $[0,100]^2$. The dialogue ends when the simulator stops or
reaches $T_{\max}$. Before producing the state update, the simulator first writes a brief
internal explanation based on the dialogue history. 
Full prompts are in Appendix ~\ref{app:prompt}.

Each scenario has start, successful, and failed anchors on both axes:
$(a_{\mathrm{start}},a_{\mathrm{succ}},a_{\mathrm{fail}})$ and
$(t_{\mathrm{start}},t_{\mathrm{succ}},t_{\mathrm{fail}})$.
The start anchor gives score $0$, the successful anchor gives $+1$,
and the failed anchor gives $-1$.
This normalization makes different scenarios comparable: each final state is scored by its progress toward the scenario-specific successful or failed state, rather than by an absolute state value.
The branch is chosen by whether $x$ moves from the start anchor toward
the successful anchor or the failed anchor.
For any axis value $x$ with anchors
$(x_{\mathrm{start}},x_{\mathrm{succ}},x_{\mathrm{fail}})$, we compute
\begin{equation}
\label{eq:piecewise}
f(x; x_{\mathrm{start}}, x_{\mathrm{succ}}, x_{\mathrm{fail}})
=
\operatorname{clip}_{[-1,+1]}
\begin{cases}
\dfrac{x - x_{\mathrm{start}}}{x_{\mathrm{succ}} - x_{\mathrm{start}}}, 
& \text{toward } x_{\mathrm{succ}}, \\[6pt]
-\dfrac{x - x_{\mathrm{start}}}{x_{\mathrm{fail}} - x_{\mathrm{start}}}, 
& \text{toward } x_{\mathrm{fail}}.
\end{cases}
\end{equation}
Thus, $f(x)=0$ at the start anchor, $f(x)=+1$ at the successful anchor, and $f(x)=-1$ at the failed anchor, with values clipped when the final state goes beyond these anchors.

Let $s_{\mathrm{final}}=(a_{\mathrm{final}},t_{\mathrm{final}})$ be the
final simulator state.
The final score, denoted as $r^{\mathrm{out}}$ for training, is the
weighted average of the two axis scores:
\begin{equation}
\label{eq:outcome-reward}
r^{\mathrm{out}} =
\lambda\, f(t_{\mathrm{final}}; t_{\mathrm{start}}, t_{\mathrm{succ}}, t_{\mathrm{fail}})
+ (1 - \lambda)\, f(a_{\mathrm{final}}; a_{\mathrm{start}}, a_{\mathrm{succ}}, a_{\mathrm{fail}}).
\end{equation}
Evaluation uses only this final score.
During RL training, we also use the intermediate states as turn-level
feedback, without extra judge-model calls.

\section{Centered Turn-Credit Shaping}
\label{sec:method}

Standard GRPO assigns the same trajectory-level credit to all turns in a rollout based on the final outcome reward.
This provides limited credit assignment in multi-turn emotional dialogue: a rollout may contain turns that improve the user state and turns that worsen it, but the final reward cannot identify which turns caused these changes.
\benchname{} provides an additional signal, since the simulator updates the user's emotion and relational state after each model reply.

\subsection{Dialogue Rollouts and Reward Signals}

We model each \benchname{} scenario as a finite-horizon dialogue between the
policy $\pi_\theta$ and the simulator.
At turn $\tau$, the policy observes the dialogue history $h_{\tau-1}$ and generates a reply
$y_\tau \sim \pi_\theta(\cdot \mid h_{\tau-1})$.
The simulator then responds with $u_\tau$ and updates the user state to
$s_\tau=(a_\tau,t_\tau)$, where $a_\tau$ is negative emotion intensity
and $t_\tau$ is relational state.
The dialogue ends when the simulator stops or when the maximum turn limit is reached.

Each scenario has fixed anchors
$(s_{\mathrm{start}},s_{\mathrm{succ}},s_{\mathrm{fail}})$.
As defined in Section~\ref{sec:bench:simulator}, the final state
$s_{\mathrm{final}}$ is mapped to an outcome reward
$r^{\mathrm{out}}\in[-1,1]$.
This reward measures the overall result of the dialogue and is the only reward used for evaluation.
During training, we also use the simulator's per-turn state updates.
Let $\Delta a_\tau=a_\tau-a_{\tau-1}$ and
$\Delta t_\tau=t_\tau-t_{\tau-1}$.
We define a process reward for each turn:
\begin{equation}
r^{\mathrm{proc}}_\tau
=
\frac{1}{2}\cdot\frac{-\Delta a_\tau}{100}
+
\frac{1}{2}\cdot\frac{\Delta t_\tau}{100}.
\end{equation}
This process reward measures local improvement: it rewards turns that make the user less upset or more willing to engage, regardless of the final outcome of the whole dialogue.
The factor $100$ normalizes both axes, since
$a_\tau,t_\tau\in[0,100]$.

\subsection{Centered Turn-Credit Objective}
\label{sec:method:reshape}

For each prompt $p$, we sample $K$ rollouts under the same scenario
anchors. The trajectory-level advantage follows standard GRPO
group-normalization:
\begin{equation}
\label{eq:grpo-adv}
A^{\mathrm{trace},(k)}
=
\frac{r^{\mathrm{out},(k)} - \mu_p}
{\max(\sigma_p,\,\sigma_{\min})},
\end{equation}
where $\mu_p$ and $\sigma_p$ are the mean and standard deviation of
$\{r^{\mathrm{out},(k)}\}_{k=1}^K$. We apply a small lower bound $\sigma_{\min}$ to the group standard deviation to avoid overly large advantages when rollout scores are nearly identical.
A natural baseline is to add the process reward directly to the
trajectory-level advantage:
\begin{equation}
\label{eq:naiveadv}
A^{(k)}_\tau
=
A^{\mathrm{trace},(k)}
+
\alpha \, r^{\mathrm{proc},(k)}_\tau .
\end{equation}
However, direct addition changes the overall rollout advantage, not only
the relative credit of different turns. To see this, let
$\bar{r}^{\mathrm{proc},(k)}
=
\tfrac{1}{T^{(k)}} \sum_{\tau'=1}^{T^{(k)}}
r^{\mathrm{proc},(k)}_{\tau'}$
be the mean process reward of rollout $k$. Then Eq.~\ref{eq:naiveadv}
can be written as
\begin{equation}
\label{eq:decomp}
A^{(k)}_\tau =
\underbrace{\bigl(A^{\mathrm{trace},(k)} + \alpha\,\bar{r}^{\mathrm{proc},(k)}\bigr)}_{\text{shared offset}}
+ \underbrace{\alpha\bigl(r^{\mathrm{proc},(k)}_\tau - \bar{r}^{\mathrm{proc},(k)}\bigr)}_{\text{turn-specific deviation}}.
\end{equation}
The shared offset is applied to all turns in the rollout, so it acts as
an extra rollout-level reward after GRPO group normalization. This changes
the normalized outcome advantage, while only the deviation term gives
turn-specific credit.
We therefore center the process reward within each rollout:
\begin{equation}
\label{eq:turncredit}
A^{(k)}_\tau
=
A^{\mathrm{trace},(k)}
+
\alpha
\bigl(
r^{\mathrm{proc},(k)}_\tau
-
\bar{r}^{\mathrm{proc},(k)}
\bigr).
\end{equation}
The centered term sums to zero across turns in the same rollout. Thus,
$A^{\mathrm{trace},(k)}$ keeps the outcome-level signal, while the
centered process reward only redistributes credit among turns.
This gives turn-specific credit without changing the rollout-level outcome signal.

The final training objective is the standard GRPO clipped objective with
$A^{(k)}_\tau$ from Eq.~\ref{eq:turncredit}. Let
$\rho^{(k)}_{\tau,i}(\theta)$ be the token-level importance ratio at the
$i$-th token of turn $\tau$ in rollout $k$. The loss is

\begin{equation}
\label{eq:loss}
\mathcal{L}(\theta) = -\mathbb{E}\bigl[\min\bigl(\rho^{(k)}_{\tau,i}(\theta) A^{(k)}_\tau,\, \operatorname{clip}(\rho^{(k)}_{\tau,i}(\theta), 1-\epsilon, 1+\epsilon) A^{(k)}_\tau\bigr)\bigr] + \beta\, \mathrm{KL}(\pi_\theta \Vert \pi_{\mathrm{ref}}).
\end{equation}
where $\epsilon$ is the clipping threshold and $\beta$ is the coefficient
of the KL penalty against the fixed $\pi_{\mathrm{ref}}$.

\section{Experiments}
\label{sec:exp}

\begin{table*}[t]
\centering
\small
\setlength{\tabcolsep}{3.5pt}
\begin{tabular}{l ccccc c c c c c}
\toprule
& \multicolumn{6}{c}{Qwen3-Max} & \multicolumn{2}{c}{DS-V4-Pro} & \multicolumn{2}{c}{Arena} \\
\cmidrule(lr){2-7}\cmidrule(lr){8-9}\cmidrule(l){10-11}
Model & \scCharm{} & \scDefense{} & \scRepair{} & \scSupport{} & Avg & Rank & Avg & Rank & Elo & Rank \\
\midrule
Claude-Sonnet-4.6      & $+43.8$ & $-14.6$ & $+17.9$ & $+56.5$ & $+23.0$ & $1$  & $+24.0$ & $1$  & $1548$ & $4$  \\
GPT-5.4                & $+41.5$ &  $-9.5$ & $+14.5$ & $+49.0$ & $+21.4$ & $2$  & $+22.7$ & $3$  & $1593$ & $3$  \\
Qwen3-Max              & $+32.0$ &  $-2.4$ & $+14.1$ & $+45.7$ & $+20.6$ & $3$  & $+14.8$ & $11$ & $1487$ & $8$  \\
Gemini-3.1-Pro         & $+31.3$ &  $-2.5$ & $+18.6$ & $+39.3$ & $+20.0$ & $4$  & $+21.5$ & $5$  & $1616$ & $1$  \\
Kimi-K2.6              & $+37.3$ &  $-8.8$ & $+17.6$ & $+42.1$ & $+19.8$ & $5$  & $+21.9$ & $4$  & $1478$ & $10$ \\
GLM-5.1                & $+37.9$ & $-15.2$ & $+19.3$ & $+44.8$ & $+19.1$ & $6$  & $+19.4$ & $7$  & $1504$ & $7$  \\
Qwen3.6-Max-Preview    & $+39.1$ & $-14.3$ & $+14.7$ & $+42.7$ & $+17.9$ & $7$  & $+19.9$ & $6$  & $1519$ & $6$  \\
Gemini-3-Flash         & $+37.0$ & $-11.1$ & $+18.1$ & $+34.0$ & $+17.2$ & $8$  & $+16.5$ & $10$ & $1606$ & $2$  \\
Doubao-Seed-2.0-Pro    & $+31.9$ & $-14.2$ & $+10.7$ & $+39.8$ & $+14.8$ & $9$  & $+17.0$ & $9$  & $1536$ & $5$  \\
DeepSeek-V4-Pro        & $+28.0$ &  $-7.8$ & $+12.2$ & $+32.4$ & $+14.4$ & $10$ & $+23.3$ & $2$  & $1484$ & $9$  \\
Grok-4.20-Reasoning    & $+28.0$ & $-19.0$ & $+12.2$ & $+40.9$ & $+13.1$ & $11$ & $+10.6$ & $13$ & $1458$ & $13$ \\
Grok-4.20              & $+31.0$ & $-13.5$ &  $+5.1$ & $+38.4$ & $+13.1$ & $12$ & $+13.0$ & $12$ & $1476$ & $11$ \\
MiniMax-M2.5           & $+21.6$ & $-16.7$ &  $+5.2$ & $+31.4$ &  $+8.4$ & $13$ & $+17.6$ & $8$  & $1471$ & $12$ \\
Qwen3-32B              & $+23.7$ & $-14.3$ &  $+6.8$ &  $+3.9$ &  $+3.4$ & $14$ &  $+5.8$ & $14$ & $1405$ & $14$ \\
Qwen3-8B               & $+14.6$ & $-21.4$ & $-35.9$ & $-43.3$ & $-22.4$ & $15$ & $-23.7$ & $15$ & $1316$ & $15$ \\
\bottomrule
\end{tabular}
\caption{\benchname{} leaderboard of $15$ LLMs. For Qwen3-Max simulator we report per-scene score and the weighted average; For DS-V4-Pro simulator and the human arena we report the summary score and rank only.}
\label{tab:main-leaderboard}
\end{table*}

\subsection{Setup}
We evaluate 15 open- and closed-source LLMs on EIBench using
Qwen3-Max as the main simulator and DeepSeek-V4-Pro as another
simulator from a different model family. 
Both simulators use the same
per-scenario \texttt{simulated\_user\_profile} and state anchors. 
We collect a human pairwise arena with more than 4,000 annotations to
validate the simulator rankings.

For RL, we train on the EIBench training split with Qwen3-Max as the
simulator and verl as the training framework. Unless otherwise stated,
we use $T_{\max}=8$, the outcome reward in Eq.~\ref{eq:outcome-reward} with
$\lambda=0.5$, turn-credit weight $\alpha=15$, and GRPO normalization
floor $\sigma_{\min}=0.1$. Additional training details, hyperparameters, and cost estimates are provided in Appendix~\ref{app:hyperparams}.
We hold out the EIBench test split and also
evaluate on SAGE and EQBench3 under their default settings.

\subsection{Leaderboard}
\label{sec:exp:leaderboard}

Table~\ref{tab:main-leaderboard} reports per-scene and average scores under the
Qwen3-Max simulator.
A positive score means that the final state is better than the initial state,
with lower negative emotion and higher relational state; a negative score means
that the interaction moves the user state in the wrong direction.
The main pattern is clear: current LLMs are much stronger at comfort and rapport
building than at boundary management.
Most models obtain positive scores on \scSupport{}, \scCharm{}, and
\scRepair{}, but all models score below zero on \scDefense{}.

\textbf{\scDefense{} reveals a boundary-management weakness.}
All $15$ models score below zero on \scDefense{}.
This means the simulated user ends in a worse state than where the dialogue began.
This scene requires the model to resist user pressure while also calming the user and keeping the conversation open.
In sampled dialogues, models often handle this balance poorly.
Some keep the boundary in a rigid way, with little comfort or practical explanation.
Others produce long, template-like responses that sound polite but do not directly address the user's concern.
As a result, the user often remains frustrated.
Excluding the Qwen3-Max simulator's own model because of self-bias, only Gemini-3.1-Pro ($-2.5$) roughly keeps the user's state stable, but it still does not improve it.

\textbf{The easier scenes align more closely with standard
post-training behavior.}
Most models perform best on \scSupport{} and \scCharm{}, where the model can improve the interaction mainly by being warm, encouraging, and attentive.
These behaviors are close to the helpful and supportive style encouraged by common post-training.
By contrast, \scDefense{} requires the model to combine warmth with constraint: it must acknowledge the user's emotion, keep a necessary boundary, and still offer a useful next step.
This contrast suggests that good performance on supportive dialogue does not necessarily imply broad emotion-management ability.

\begin{figure}[t]
\centering
\includegraphics[width=0.6\linewidth]{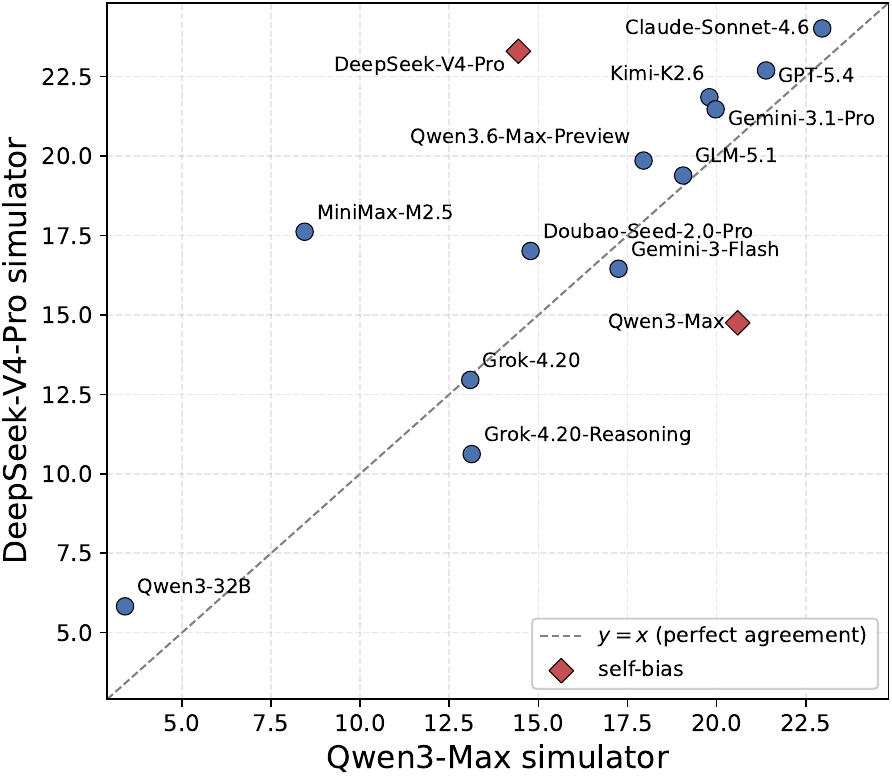}
\caption{\benchname{} scores from the Qwen3-Max and DS-V4-Pro simulators. Each point is one model. Most models are close to the diagonal; the red diamonds mark the simulators' own models and show self-bias.}
\label{fig:cross-sim-scatter}
\end{figure}

\subsection{Cross-Simulator Consistency}
\label{sec:exp:consistency}

To reduce dependence on a single simulator, we use DeepSeek-V4-Pro (DS-V4-Pro) as a second simulator from a different model family.
Figure~\ref{fig:cross-sim-scatter} shows that both simulators exhibit visible self-bias: each gives a higher score to its own model.
After removing the corresponding self-scored models, the two simulator rankings are strongly aligned, with Spearman's $\rho=0.93$ (Table~\ref{rho_tab}).
This suggests that the EIBench ranking is reasonably robust across different simulators.

We further validate EIBench against human judgment with a pairwise arena.
Annotators interact with two anonymized candidate models under the same scenario profile and choose which model handles the scenario better.
The human ranking is also consistent with the simulator rankings, with Spearman's $\rho=0.82$ against Qwen3-Max and $\rho=0.70$ against DS-V4-Pro.
The closer agreement with Qwen3-Max supports our choice of Qwen3-Max as the main training simulator. Full human-arena details are provided in Appendix~\ref{app:human_eval}.

\begin{table}[h]
\centering
\small
\setlength{\tabcolsep}{6pt}
\begin{tabular}{lccc}
\toprule
& Qwen3-Max & DS-V4-Pro & Arena \\
\midrule
Qwen3-Max & 1.00 & -- & -- \\
DS-V4-Pro & 0.93 & 1.00 & -- \\
Arena & 0.82 & 0.70 & 1.00 \\
\bottomrule
\end{tabular}
\caption{Rank correlation between EIBench rankings under the two simulators and the human arena.
We report Spearman's $\rho$ after excluding the corresponding self-scored model from each pair.}
\label{rho_tab}
\end{table}

The remaining human--simulator gap is concentrated on a few models.
Gemini-3.1-Pro and Gemini-3-Flash rise from mid-tier under both simulators to ranks 1 and 2 in the human arena, while Claude-Sonnet-4.6 and Kimi-K2.6 drop several places.
Inspecting sampled dialogues suggests that human annotators are more sensitive to how the model positions itself in the conversation.
Gemini often uses inclusive framings such as ``let's figure this out together,'' which places the model and the user on the same side.
Claude more often speaks as an outside helper, and Kimi-K2.6 sometimes emits inner monologue mid-reply, which several annotators found unnatural.
These patterns suggest that the simulators capture the main ranking trend, but may give less weight to some interaction cues that human annotators value.

\subsection{Main Results}
\label{sec:exp:main}

We report results under the training simulator, an unseen simulator, and two external benchmarks: Sage and EQbench3.
This design tests whether \methodname{} improves in-domain performance, transfers across simulators, and generalizes beyond EIBench to different task formats and judges.

\textbf{Targeted RL improves both in-domain and cross-simulator
performance.}
As shown in Table~\ref{tab:method-main}, on in-domain \benchname{}-Qwen3-Max, the trained 8B improves from
$-22.4$ to $+22.4$, and the trained 32B further reaches $+28.7$, above
every evaluated closed-source system under this simulator.
This shows that \methodname{} can effectively use the training simulator's
feedback and improve in-domain performance.
The gains also transfer to the unseen DS-V4-Pro simulator: the trained 8B
and 32B reach $+16.6$ and $+18.9$, well above their bases ($-23.7$ and
$+5.8$).
The cross-simulator drop is not uniform.
It is larger on \scCharm{} ($-8.2$ for the 8B) and \scSupport{} ($-9.1$),
and smaller on \scRepair{} and \scDefense{}.
One possible reason is that \scCharm{} and \scSupport{} depend more on
subjective judgments of warmth and comfort, which the two simulators may
weight differently.
By contrast, \scRepair{} and \scDefense{} involve more concrete behaviors
such as acknowledgment and boundary holding.
This suggests that \methodname{} learns partly transferable interaction
strategies, while still reflecting the preferences of the training
simulator.

\textbf{Gains hold on two out-of-distribution benchmarks.} On SAGE, a support-only benchmark with a DeepSeek-V3 simulator, both
model sizes improve by roughly $+12$ over their bases. On EQBench3, a
broader social-reasoning benchmark judged by Claude-Opus-4.6, the
trained 32B reaches rubric $84.5$ and Elo $1150$, ranking $\#6$ by
rubric and $\#20$ by Elo on the public leaderboard. Since EQBench3
differs from our training setup in judge, task format, and task scope,
these results suggest that the gains are not only capturing the
preferences of the Qwen3-Max simulator.

\begin{table*}[t]
\centering
\small
\setlength{\tabcolsep}{6pt}
\begin{tabular}{lccccc}
\toprule
& \multicolumn{2}{c}{\benchname{}} & \multicolumn{3}{c}{External} \\
\cmidrule(lr){2-3}\cmidrule(lr){4-6}
Model & Qwen3-Max & DeepSeek-v4-pro & SAGE & EQBench3 rubric& EQBench3 Elo \\
\midrule
Qwen3-8B base  & $-22.4$ & $-23.7$ & $16.3$ & $73.5$ (\#32) & $745$ (\#38) \\
Qwen3-32B base & $+3.4$  & $+5.8$  & $27.5$ & $74.3$ (\#33) & $948$ (\#31) \\
\midrule
\methodname{} 8B  & $\mathbf{+22.4}$  & $\mathbf{+16.6}$  & $\mathbf{28.7}$ & $\mathbf{80.1}$ (\#22) & $\mathbf{901}$ (\#31) \\
\methodname{} 32B & $\mathbf{+28.7}$      & $\mathbf{+18.9}$      & $\mathbf{39.6}$ & $\mathbf{84.5}$ (\#6)   & $\mathbf{1150}$ (\#20) \\
\bottomrule
\end{tabular}
\caption{Main results of \methodname{} on \benchname{} and external benchmarks. Numbers in parentheses indicate public ranks. EQBench3 base-model scores and public ranks are taken from its GitHub repository.}
\label{tab:method-main}
\end{table*}

\begin{figure*}[h]
\centering
\includegraphics[width=0.75\linewidth]{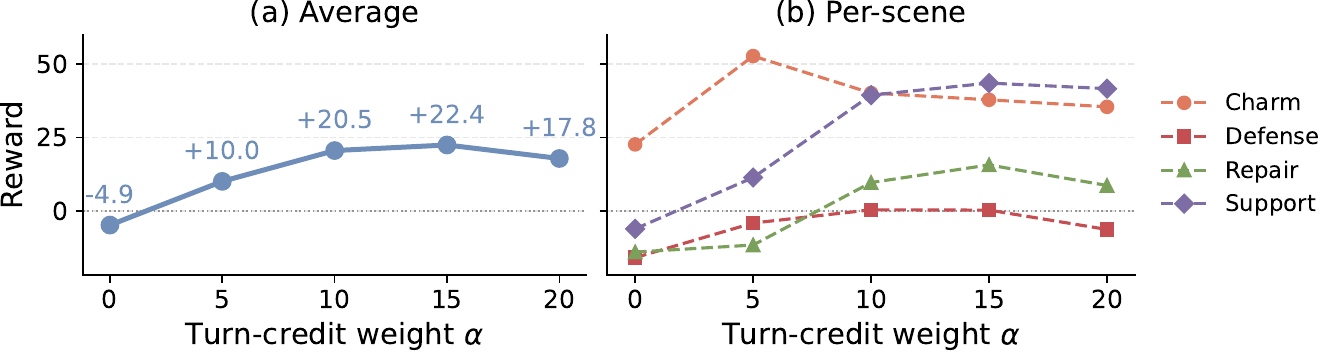}
\caption{Effect of the turn-credit weight $\alpha$ on \benchname{}: average and per-scene reward.}
\label{fig:alpha-sweep}
\end{figure*}

\subsection{Analysis}
\label{sec:exp:analysis}

\subsubsection{Hyperparameter Sensitivity}
\label{sec:exp:sweep}

The turn-credit weight $\alpha$ controls how much the centered process
reward affects the trajectory advantage. The two terms are on different
scales: $A^{\mathrm{trace}}$ is order $1$, while
$r^{\mathrm{proc}}_\tau - \bar{r}^{\mathrm{proc}}$ is usually order
$10^{-2}$. Therefore, $\alpha$ should be around $10$ for the process
reward to have a comparable effect.
Figure~\ref{fig:alpha-sweep} tests
$\alpha \in \{0,5,10,15,20\}$. With $\alpha=0$, the method reduces to
vanilla GRPO with only the trajectory reward, and the average score
remains low. The average improves as $\alpha$ increases to $10$--$15$,
then drops at $\alpha=20$, suggesting that too large a turn-credit
weight may overweight the process signal. The per-scene curves show the
same pattern: moderate values improve several scenes at once, while
smaller or larger values give a less balanced trade-off. We therefore
use $\alpha=15$ as the default.

\begin{figure*}[!t]
\centering
\includegraphics[width=0.75\linewidth]{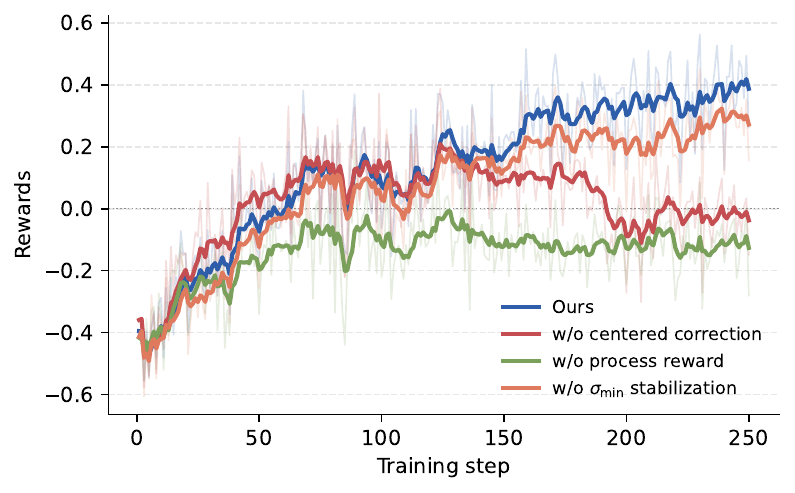}
\caption{Training reward over steps for the four ablation runs of
Table~\ref{tab:ablation} (8B, \benchname{} training split, Qwen3-Max
simulator). Curves are EMA-smoothed ($\beta{=}0.8$); the raw signal
is shown faintly behind each line.}
\label{fig:training-curves}
\end{figure*}
\subsubsection{Ablation Studies}
\label{sec:exp:ablation}

We ablate three components of \methodname{} on Qwen3-8B: the per-turn
process reward, the centered correction, and the $\sigma_{\min}$ floor
in GRPO normalization.
Table~\ref{tab:ablation} reports the final evaluation scores, and
Figure~\ref{fig:training-curves} shows the corresponding training curves.

\begin{table}[h]
\centering\small
\setlength{\tabcolsep}{3pt}
\begin{tabular}{@{}l c c c c c@{}}
\toprule
                    & Charm & Defense & Repair & Support & Avg. \\
\midrule
\methodname{}       & $+37.8$ & $+0.2$  & $+15.6$ & $+43.5$ & $\mathbf{+22.4}$ \\
\midrule
w/o centering       & $+14.2$ & $-2.0$  & $+7.8$  & $+23.6$ & $+10.0$ \\
w/o process         & $+22.6$ & $-16.0$ & $-14.1$ & $-6.2$  & $-4.9$ \\
w/o $\sigma_{\min}$ & $+37.2$ & $-11.1$ & $+10.1$ & $+30.2$ & $+14.4$ \\
\bottomrule
\end{tabular}
\caption{Ablation results on \benchname{}.}
\label{tab:ablation}
\end{table}

\textbf{Per-turn feedback is the largest contributor.}
Removing the process reward (w/o process) drops the average score from
$+22.4$ to $-4.9$, with \scDefense{} and \scRepair{} becoming negative
($-16.0$ and $-14.1$).
The training curve also shows that this variant stops improving after about
80 steps and then fluctuates around a low reward level.
These results suggest that terminal-only training provides too sparse a
learning signal for multi-turn emotional dialogue: the policy receives only a
delayed outcome reward and cannot easily identify which replies improved or
worsened the user state.
By contrast, the denser turn-level feedback helps the policy continue improving
after the early stage.

\textbf{Centering isolates turn-level credit.}
The uncentered variant (w/o centering) reaches only $+10.0$, far below
the full method.
Its training curve initially improves, but starts to decline shortly after
the reward becomes positive.
This is consistent with the analysis in \S\ref{sec:method:reshape}: once
rollouts begin to improve, the mean process reward of a rollout can become
positive, and without centering this mean is added to all turns as a shared
offset.
This extra rollout-level signal can interfere with the normalized outcome
advantage, rather than only redistributing credit across turns.
Centering removes this offset, so the process reward mainly provides
turn-level credit while keeping the outcome-level signal unchanged.

\textbf{The $\sigma_{\min}$ floor improves stability.}
Removing $\sigma_{\min}$ (w/o $\sigma_{\min}$) lowers the average score
to $+14.4$.
Among the three ablations, its training curve is the closest to the full method,
but it still ends at a lower reward.
In GRPO-style group normalization, the outcome advantage is divided by the
within-group reward standard deviation.
When this standard deviation becomes very small, small differences between
rollout rewards can be over-amplified into large advantages.
We observe this issue on \scDefense{} and \scSupport{} at some training steps.
The $\sigma_{\min}$ floor limits this over-amplification, so the update is less
sensitive to near-tied rollout scores.
This helps preserve useful trajectory-level learning while avoiding unstable
late-stage updates.

\begin{table}[t]
\centering
\small
\setlength{\tabcolsep}{5pt}
\begin{tabular}{lcccc}
\toprule
Model & AIME & GPQA & HE+ & IFEval \\
\midrule
Qwen3-8B base & 70.0 & 59.1 & 86.0 & 90.4 \\

+ \methodname{} & 67.9 & 58.3 & 87.8 & 90.1 \\
  \midrule
  Qwen3-32B base & 73.3 & 65.5 & 89.0 & 89.9 \\
+ \methodname{} & 69.6 & 65.2 & 89.0 & 89.1 \\
  \bottomrule
  \end{tabular}
  \caption{General-capability evaluation before and after \methodname{} training. We report AIME 2025, GPQA Diamond, HumanEval+ (HE+), and IFEval instruction-level strict accuracy. Higher is better.}
  \label{tab:general-capability}
  \end{table}


\subsubsection{IQ vs.\ EQ: Does EQ Training Hurt General Capability?}
\label{sec:exp:iq-vs-eq}

A natural concern with EQ-targeted RL is whether it harms general capability.
We evaluate the trained 8B and 32B models on four standard benchmarks covering math, hard scientific reasoning, code, and instruction following: AIME 2025\citep{AIME}, GPQA Diamond \citep{gpqa}, HumanEval+ \citep{he+}, and IFEval \citep{ife}.
Table~\ref{tab:general-capability} reports the results, with evaluation details provided in Appendix~\ref{app:iq-vs-eq}.

Overall, \methodname{} has little effect on general capability.
The trained models remain close to their bases on GPQA Diamond, HumanEval+, and IFEval, and HumanEval+ slightly improves for the 8B model.
AIME 2025 shows a small decrease at both scales, suggesting that EQ-targeted training may slightly affect difficult math reasoning.
However, there is no broad degradation across the benchmark suite.
These results suggest that \methodname{} largely preserves the base model's general capability profile while improving emotional dialogue performance.

\section{Conclusion}

In this work, we presented \benchname{}, a simulator-based benchmark for multi-turn emotion management, and \methodname{}, which uses simulator-tracked state changes for turn-level credit assignment.
\benchname{} evaluates whether a model can improve a simulated user's emotional and relational state across support, defense, repair, and charm scenarios.
The leaderboard shows uneven emotion-management ability in current LLMs: they are stronger at support and rapport building, but still struggle with boundary maintenance under user pressure.
\methodname{} substantially improves model performance on \benchname{} and generalizes to external emotional and social evaluations, while largely preserving performance on general benchmarks.
Overall, simulator-tracked user states provide a useful signal for both evaluation and training in multi-turn emotion management.

\section*{Acknowledgements}

This work was supported by the Alibaba Research Intern Program and the Fundamental and Interdisciplinary Disciplines Breakthrough Plan of the Ministry of Education of China (No.\ JYB2025XDXM118).

\section*{Limitations}

Our scoring uses an LLM simulator, so the per-turn state updates are
not exact and the reward signal is noisy. Cross-simulator agreement
and the human arena show that this noise does not change the overall
ranking, but it is still there. The four scenes are also not equally
hard. \scDefense{} is the hardest one, and even top closed-source
models stay slightly below zero on it. We train all four scenes with
the same hyperparameters and a single turn-credit weight $\alpha$,
and methods that handle each scene differently, such as a different
$\alpha$ per scene or different sampling rates, are left for future
work.

\clearpage
\bibliography{references}

@article{ei,
  title={What is emotional intelligence},
  author={Mayer, John D and Salovey, Peter and others},
  journal={Emotional development and emotional intelligence: Educational implications},
  volume={3},
  number={31},
  pages={1--16},
  year={1997}
}

@article{gemini,
  author       = {Gemini Team},
  title        = {Gemini 2.5: Pushing the Frontier with Advanced Reasoning, Multimodality,
                  Long Context, and Next Generation Agentic Capabilities},
  journal      = {CoRR},
  volume       = {abs/2507.06261},
  year         = {2025},
  doi          = {10.48550/ARXIV.2507.06261},
  eprinttype   = {arXiv},
  eprint       = {2507.06261},
  bibsource    = {dblp computer science bibliography, https://dblp.org}
}

@book{emotion,
  title={Emotion and adaptation},
  author={Lazarus, Richard S},
  year={1991},
  publisher={Oxford University Press}
}

@article{rlver,
  title={RLVER: REINFORCEMENT LEARNING WITH VERIFI-ABLE EMOTION REWARDS FOR EMPATHETIC AGENTS},
  author={Wang, Peisong and Ma, Ruotian and Zhang, Bang and Chen, Xingyu and He, Zhiwei and Luo, Kang and Lv, Qingsong and Jiang, Qingxuan and Xie, Zheng and Wang, Shanyi and others},
journal={arXiv preprint arXiv:2507.03112},
year = {2025}

}

@article{mapo,
  title={MAPO: Mixed Advantage Policy Optimization for Long-Horizon Multi-Turn Dialogue},
  author={Zhang, Naifan and Sun, Ruihan and Su, Jinwei and Yang, Hengjie and Pan, Zhengyuan and Chen, Zhaohan and Zhang, Xiaofan},
  journal={arXiv preprint arXiv:2603.06194},
  year={2026}
}

@article{eqbench,
  title={Eq-bench: An emotional intelligence benchmark for large language models},
  author={Paech, Samuel J},
  journal={arXiv preprint arXiv:2312.06281},
  year={2023}
}

@article{sage,
  title={Sentient agent as a judge: Evaluating higher-order social cognition in large language models},
  author={Zhang, Bang and Ma, Ruotian and Jiang, Qingxuan and Wang, Peisong and Chen, Jiaqi and Xie, Zheng and Chen, Xingyu and Wang, Yue and Ye, Fanghua and Li, Jian and others},
  journal={arXiv preprint arXiv:2505.02847},
  year={2025}
}

@article{echo,
  title={Echo-n1: Affective rl frontier},
  author={Zhang, Naifan and Sun, Ruihan and Su, Ruixi and Ma, Shiqi and Zhang, Shiya and Weng, Xianna and Zhang, Xiaofan and Zhan, Yuhan and Xu, Yuyang and Chen, Zhaohan and others},
  journal={arXiv preprint arXiv:2512.00344},
  year={2025}
}

@article{deepseekmath,
  author       = {Zhihong Shao and
                  Yuxiang Luo and
                  Chengda Lu and
                  Z. Z. Ren and
                  Jiewen Hu and
                  Tian Ye and
                  Zhibin Gou and
                  Shirong Ma and
                  Xiaokang Zhang},
  title        = {DeepSeekMath-V2: Towards Self-Verifiable Mathematical Reasoning},
  journal      = {CoRR},
  volume       = {abs/2511.22570},
  year         = {2025},
  doi          = {10.48550/ARXIV.2511.22570},
  eprinttype   = {arXiv},
  eprint       = {2511.22570},
  bibsource    = {dblp computer science bibliography, https://dblp.org}
}

@article{dapo,
  title={Dapo: An open-source llm reinforcement learning system at scale},
  author={Yu, Qiying and Zhang, Zheng and Zhu, Ruofei and Yuan, Yufeng and Zuo, Xiaochen and Yue, Yu and Dai, Weinan and Fan, Tiantian and Liu, Gaohong and Liu, Lingjun and others},
  journal={Advances in Neural Information Processing Systems},
  volume={38},
  pages={113222--113244},
  year={2026}
}

@article{vapo,
  title={Vapo: Efficient and reliable reinforcement learning for advanced reasoning tasks},
  author={Yue, Yu and Yuan, Yufeng and Yu, Qiying and Zuo, Xiaochen and Zhu, Ruofei and Xu, Wenyuan and Chen, Jiaze and Wang, Chengyi and Fan, TianTian and Du, Zhengyin and others},
  journal={arXiv preprint arXiv:2504.05118},
  year={2025}
}

@article{searchr1,
  title={Search-r1: Training llms to reason and leverage search engines with reinforcement learning},
  author={Jin, Bowen and Zeng, Hansi and Yue, Zhenrui and Yoon, Jinsung and Arik, Sercan and Wang, Dong and Zamani, Hamed and Han, Jiawei},
  journal={arXiv preprint arXiv:2503.09516},
  year={2025}
}

@article{gigpo,
  title={Group-in-group policy optimization for llm agent training},
  author={Feng, Lang and Xue, Zhenghai and Liu, Tingcong and An, Bo},
  journal={Advances in Neural Information Processing Systems},
  volume={38},
  pages={46375--46408},
  year={2026}
}

@article{qwen3coder,
  title={Qwen3-coder-next technical report},
  author={Cao, Ruisheng and Chen, Mouxiang and Chen, Jiawei and Cui, Zeyu and Feng, Yunlong and Hui, Binyuan and Jing, Yuheng and Li, Kaixin and Li, Mingze and Lin, Junyang and others},
  journal={arXiv preprint arXiv:2603.00729},
  year={2026}
}

@article{deepseekcoder,
  title={DeepSeek-Coder: when the large language model meets programming--the rise of code intelligence},
  author={Guo, Daya and Zhu, Qihao and Yang, Dejian and Xie, Zhenda and Dong, Kai and Zhang, Wentao and Chen, Guanting and Bi, Xiao and Wu, Yifan and Li, YK and others},
  journal={arXiv preprint arXiv:2401.14196},
  year={2024}
}

@article{deepseek,
  author       = {Daya Guo and
                  Dejian Yang and
                  Haowei Zhang and
                  Junxiao Song and
                  Peiyi Wang and
                  Qihao Zhu and
                  Runxin Xu and
                  Ruoyu Zhang and
                  Shirong Ma and
                  Xiao Bi and
                  Xiaokang Zhang and
                  Xingkai Yu and
                  Yu Wu and
                  Z. F. Wu and
                  Zhibin Gou and
                  Zhihong Shao and
                  Zhuoshu Li and
                  Ziyi Gao and
                  Aixin Liu and
                  Bing Xue and
                  Bingxuan Wang and
                  Bochao Wu and
                  Bei Feng and
                  Chengda Lu and
                  Chenggang Zhao and
                  Chengqi Deng and
                  Chong Ruan and
                  Damai Dai and
                  Deli Chen and
                  Dongjie Ji and
                  Erhang Li and
                  Fangyun Lin and
                  Fucong Dai and
                  Fuli Luo and
                  Guangbo Hao and
                  Guanting Chen and
                  Guowei Li and
                  Hao Zhang and
                  Hanwei Xu and
                  Honghui Ding and
                  Huazuo Gao and
                  Hui Qu and
                  Hui Li and
                  Jianzhong Guo and
                  Jiashi Li and
                  Jingchang Chen and
                  Jingyang Yuan and
                  Jinhao Tu and
                  Junjie Qiu and
                  Junlong Li and
                  J. L. Cai and
                  Jiaqi Ni and
                  Jian Liang and
                  Jin Chen and
                  Kai Dong and
                  Kai Hu and
                  Kaichao You and
                  Kaige Gao and
                  Kang Guan and
                  Kexin Huang and
                  Kuai Yu and
                  Lean Wang and
                  Lecong Zhang and
                  Liang Zhao and
                  Litong Wang and
                  Liyue Zhang and
                  Lei Xu and
                  Leyi Xia and
                  Mingchuan Zhang and
                  Minghua Zhang and
                  Minghui Tang and
                  Mingxu Zhou and
                  Meng Li and
                  Miaojun Wang and
                  Mingming Li and
                  Ning Tian and
                  Panpan Huang and
                  Peng Zhang and
                  Qiancheng Wang and
                  Qinyu Chen and
                  Qiushi Du and
                  Ruiqi Ge and
                  Ruisong Zhang and
                  Ruizhe Pan and
                  Runji Wang and
                  R. J. Chen and
                  R. L. Jin and
                  Ruyi Chen and
                  Shanghao Lu and
                  Shangyan Zhou and
                  Shanhuang Chen and
                  Shengfeng Ye and
                  Shiyu Wang and
                  Shuiping Yu and
                  Shunfeng Zhou and
                  Shuting Pan and
                  S. S. Li and
                  Shuang Zhou and
                  Shaoqing Wu and
                  Tao Yun and
                  Tian Pei and
                  Tianyu Sun and
                  Tao Wang and
                  Wangding Zeng and
                  Wen Liu and
                  Wenfeng Liang and
                  Wenjun Gao and
                  Wenqin Yu and
                  Wentao Zhang and
                  W. L. Xiao and
                  Wei An and
                  Xiaodong Liu and
                  Xiaohan Wang and
                  Xiaokang Chen and
                  Xiaotao Nie and
                  Xin Cheng and
                  Xin Liu and
                  Xin Xie and
                  Xingchao Liu and
                  Xinyu Yang and
                  Xinyuan Li and
                  Xuecheng Su and
                  Xuheng Lin and
                  X. Q. Li and
                  Xiangyue Jin and
                  Xiaojin Shen and
                  Xiaosha Chen and
                  Xiaowen Sun and
                  Xiaoxiang Wang and
                  Xinnan Song and
                  Xinyi Zhou and
                  Xianzu Wang and
                  Xinxia Shan and
                  Y. K. Li and
                  Y. Q. Wang and
                  Y. X. Wei and
                  Yang Zhang and
                  Yanhong Xu and
                  Yao Li and
                  Yao Zhao and
                  Yaofeng Sun and
                  Yaohui Wang and
                  Yi Yu and
                  Yichao Zhang and
                  Yifan Shi and
                  Yiliang Xiong and
                  Ying He and
                  Yishi Piao and
                  Yisong Wang and
                  Yixuan Tan and
                  Yiyang Ma and
                  Yiyuan Liu and
                  Yongqiang Guo and
                  Yuan Ou and
                  Yuduan Wang and
                  Yue Gong and
                  Yuheng Zou and
                  Yujia He and
                  Yunfan Xiong and
                  Yuxiang Luo and
                  Yuxiang You and
                  Yuxuan Liu and
                  Yuyang Zhou and
                  Y. X. Zhu and
                  Yanping Huang and
                  Yaohui Li and
                  Yi Zheng and
                  Yuchen Zhu and
                  Yunxian Ma and
                  Ying Tang and
                  Yukun Zha and
                  Yuting Yan and
                  Z. Z. Ren and
                  Zehui Ren and
                  Zhangli Sha and
                  Zhe Fu and
                  Zhean Xu and
                  Zhenda Xie and
                  Zhengyan Zhang and
                  Zhewen Hao and
                  Zhicheng Ma and
                  Zhigang Yan and
                  Zhiyu Wu and
                  Zihui Gu and
                  Zijia Zhu and
                  Zijun Liu and
                  Zilin Li and
                  Ziwei Xie and
                  Ziyang Song and
                  Zizheng Pan and
                  Zhen Huang and
                  Zhipeng Xu and
                  Zhongyu Zhang and
                  Zhen Zhang},
  title        = {DeepSeek-R1 incentivizes reasoning in LLMs through reinforcement learning},
  journal      = {Nat.},
  volume       = {645},
  number       = {8081},
  pages        = {633--638},
  year         = {2025},
  doi          = {10.1038/S41586-025-09422-Z},
  bibsource    = {dblp computer science bibliography, https://dblp.org}
}

@article{gpt4,
  author       = {OpenAI},
  title        = {{GPT-4} Technical Report},
  journal      = {CoRR},
  volume       = {abs/2303.08774},
  year         = {2023},
  doi          = {10.48550/ARXIV.2303.08774},
  eprinttype   = {arXiv},
  eprint       = {2303.08774},
  bibsource    = {dblp computer science bibliography, https://dblp.org}
}

@article{qwen3,
  author       = {Qwen Team},
  title        = {Qwen3 Technical Report},
  journal      = {CoRR},
  volume       = {abs/2505.09388},
  year         = {2025},
  doi          = {10.48550/ARXIV.2505.09388},
  eprinttype   = {arXiv},
  eprint       = {2505.09388},
  bibsource    = {dblp computer science bibliography, https://dblp.org}
}

@article{mathimo,
  author       = {Yuri Chervonyi and
                  Trieu H. Trinh and
                  Miroslav Ols{\'{a}}k and
                  Xiaomeng Yang and
                  Hoang H. Nguyen and
                  Marcelo Menegali and
                  Junehyuk Jung and
                  Junsu Kim and
                  Vikas Verma and
                  Quoc V. Le and
                  Thang Luong},
  title        = {Gold-medalist Performance in Solving Olympiad Geometry with AlphaGeometry2},
  journal      = {J. Mach. Learn. Res.},
  volume       = {26},
  pages        = {241:1--241:39},
  year         = {2025},
}

@article{tongyideepresearch,
  title={Tongyi deepresearch technical report},
  author={Team Tongyi DeepResearch and Li, Baixuan and Zhang, Bo and Zhang, Dingchu and Huang, Fei and Li, Guangyu and Chen, Guoxin and Yin, Huifeng and Wu, Jialong and Zhou, Jingren and others},
  journal={arXiv preprint arXiv:2510.24701},
  year={2025}
}

@article{rlvr,
  title={Reinforcement learning with verifiable rewards implicitly incentivizes correct reasoning in base llms},
  author={Wen, Xumeng and Liu, Zihan and Zheng, Shun and Ye, Shengyu and Wu, Zhirong and Wang, Yang and Xu, Zhijian and Liang, Xiao and Li, Junjie and Miao, Ziming and others},
  journal={arXiv preprint arXiv:2506.14245},
  year={2025}
}

@article{simplerl,
  title={Simplerl-zoo: Investigating and taming zero reinforcement learning for open base models in the wild},
  author={Zeng, Weihao and Huang, Yuzhen and Liu, Qian and Liu, Wei and He, Keqing and Ma, Zejun and He, Junxian},
  journal={arXiv preprint arXiv:2503.18892},
  year={2025}
}

@article{arenarl,
  title={ArenaRL: Scaling RL for Open-Ended Agents via Tournament-based Relative Ranking},
  author={Zhang, Qiang and Chen, Boli and Zhang, Fanrui and Ding, Ruixue and Wang, Shihang and Wang, Qiuchen and Huang, Yinfeng and Zhang, Haonan and Zhu, Rongxiang and Wang, Pengyong and others},
  journal={arXiv preprint arXiv:2601.06487},
  year={2026}
}

@article{llmjudge,
  title={Judging llm-as-a-judge with mt-bench and chatbot arena},
  author={Zheng, Lianmin and Chiang, Wei-Lin and Sheng, Ying and Zhuang, Siyuan and Wu, Zhanghao and Zhuang, Yonghao and Lin, Zi and Li, Zhuohan and Li, Dacheng and Xing, Eric and others},
  journal={Advances in neural information processing systems},
  volume={36},
  pages={46595--46623},
  year={2023}
}

@article{grm,
  title={Reward Modeling from Natural Language Human Feedback},
  author={Wang, Zongqi and Wang, Rui and Wu, Yuchuan and Yu, Yiyao and Zhang, Pinyi and Sun, Shaoning and Yang, Yujiu and Li, Yongbin},
  journal={arXiv preprint arXiv:2601.07349},
  year={2026}
}

@article{autorefine,
  title={Search and refine during think: Facilitating knowledge refinement for improved retrieval-augmented reasoning},
  author={Shi, Yaorui and Li, Sihang and Wu, Chang and Liu, Zhiyuan and Fang, Junfeng and Cai, Hengxing and Zhang, An and Wang, Xiang},
  journal={Advances in Neural Information Processing Systems},
  volume={38},
  pages={155930--155958},
  year={2026}
}

@inproceedings{towards_empathetic,
  title={Towards empathetic open-domain conversation models: A new benchmark and dataset},
  author={Rashkin, Hannah and Smith, Eric Michael and Li, Margaret and Boureau, Y-Lan},
  booktitle={Proceedings of the 57th annual meeting of the association for computational linguistics},
  pages={5370--5381},
  year={2019}
}

@inproceedings{psyqa,
  title={Psyqa: A chinese dataset for generating long counseling text for mental health support},
  author={Sun, Hao and Lin, Zhenru and Zheng, Chujie and Liu, Siyang and Huang, Minlie},
  booktitle={Findings of the association for computational linguistics: ACL-IJCNLP 2021},
  pages={1489--1503},
  year={2021}
}

@inproceedings{esconv,
  title={Towards emotional support dialog systems},
  author={Liu, Siyang and Zheng, Chujie and Demasi, Orianna and Sabour, Sahand and Li, Yu and Yu, Zhou and Jiang, Yong and Huang, Minlie},
  booktitle={Proceedings of the 59th annual meeting of the association for computational linguistics and the 11th international joint conference on natural language processing (volume 1: Long papers)},
  pages={3469--3483},
  year={2021}
}

@inproceedings{emobench,
  title={Emobench: Evaluating the emotional intelligence of large language models},
  author={Sabour, Sahand and Liu, Siyang and Zhang, Zheyuan and Liu, June and Zhou, Jinfeng and Sunaryo, Alvionna and Lee, Tatia and Mihalcea, Rada and Huang, Minlie},
  booktitle={Proceedings of the 62nd Annual Meeting of the Association for Computational Linguistics (Volume 1: Long Papers)},
  pages={5986--6004},
  year={2024}
}

@inproceedings{tombench,
  title={ToMBench: Benchmarking theory of mind in large language models},
  author={Chen, Zhuang and Wu, Jincenzi and Zhou, Jinfeng and Wen, Bosi and Bi, Guanqun and Jiang, Gongyao and Cao, Yaru and Hu, Mengting and Lai, Yunghwei and Xiong, Zexuan and others},
  booktitle={Proceedings of the 62nd Annual Meeting of the Association for Computational Linguistics (Volume 1: Long Papers)},
  pages={15959--15983},
  year={2024}
}

@article{sibench,
  title={SI-Bench: Benchmarking Social Intelligence of Large Language Models in Human-to-Human Conversations},
  author={Huang, Shuai and Zhao, Wenxuan and Gao, Jun},
  journal={arXiv preprint arXiv:2510.23182},
  year={2025}
}

@inproceedings{agentsense,
  title={AgentSense: LLMs empower generalizable and explainable web-based participatory urban sensing},
  author={Guo, Xusen and Peng, Mingxing and Hao, Xixuan and Zou, Xingchen and Wang, Qiongyan and Ruan, Sijie and Liang, Yuxuan},
  booktitle={Proceedings of the ACM Web Conference 2026},
  pages={5439--5450},
  year={2026}
}

@inproceedings{sotopia,
  title={Sotopia: Interactive evaluation for social intelligence in language agents},
  author={Zhou, Xuhui and Zhu, Hao and Mathur, Leena and Zhang, Ruohong and Yu, Haofei and Qi, Zhengyang and Morency, Louis-Philippe and Bisk, Yonatan and Fried, Daniel and Neubig, Graham and others},
  booktitle={International Conference on Learning Representations},
  volume={2024},
  pages={40975--41019},
  year={2024}
}

@article{gspo,
  title={Group sequence policy optimization},
  author={Zheng, Chujie and Liu, Shixuan and Li, Mingze and Chen, Xiong-Hui and Yu, Bowen and Gao, Chang and Dang, Kai and Liu, Yuqiong and Men, Rui and Yang, An and others},
  journal={arXiv preprint arXiv:2507.18071},
  year={2025}
}

@inproceedings{coser,
  author       = {Xintao Wang and
                  Heng Wang and
                  Yifei Zhang and
                  Xinfeng Yuan and
                  Rui Xu and
                  Jen{-}tse Huang and
                  Siyu Yuan and
                  Haoran Guo and
                  Jiangjie Chen and
                  Shuchang Zhou and
                  Wei Wang and
                  Yanghua Xiao},
  editor       = {Aarti Singh and
                  Maryam Fazel and
                  Daniel Hsu and
                  Simon Lacoste{-}Julien and
                  Felix Berkenkamp and
                  Tegan Maharaj and
                  Kiri Wagstaff and
                  Jerry Zhu},
  title        = {CoSER: Coordinating LLM-Based Persona Simulation of Established Roles},
  booktitle    = {Forty-second International Conference on Machine Learning, {ICML}
                  2025, Vancouver, BC, Canada, July 13-19, 2025},
  year         = {2025},
}

@article{gpqa,
  author       = {David Rein and
                  Betty Li Hou and
                  Asa Cooper Stickland and
                  Jackson Petty and
                  Richard Yuanzhe Pang and
                  Julien Dirani and
                  Julian Michael and
                  Samuel R. Bowman},
  title        = {{GPQA:} {A} Graduate-Level Google-Proof Q{\&}A Benchmark},
  journal      = {CoRR},
  volume       = {abs/2311.12022},
  year         = {2023},
}

@inproceedings{he+,
  author       = {Jiawei Liu and
                  Chunqiu Steven Xia and
                  Yuyao Wang and
                  Lingming Zhang},
  editor       = {Alice Oh and
                  Tristan Naumann and
                  Amir Globerson and
                  Kate Saenko and
                  Moritz Hardt and
                  Sergey Levine},
  title        = {Is Your Code Generated by ChatGPT Really Correct? Rigorous Evaluation
                  of Large Language Models for Code Generation},
  booktitle    = {Advances in Neural Information Processing Systems 36: Annual Conference
                  on Neural Information Processing Systems 2023, NeurIPS 2023, New Orleans,
                  LA, USA, December 10 - 16, 2023},
  year         = {2023},
}

@article{ife,
  author       = {Jeffrey Zhou and
                  Tianjian Lu and
                  Swaroop Mishra and
                  Siddhartha Brahma and
                  Sujoy Basu and
                  Yi Luan and
                  Denny Zhou and
                  Le Hou},
  title        = {Instruction-Following Evaluation for Large Language Models},
  journal      = {CoRR},
  volume       = {abs/2311.07911},
  year         = {2023},
}

@misc{AIME,
author       = {{OpenCompass}},
title        = {{AIME} 2025 Dataset},
year         = {2025},
howpublished = {\url{https://huggingface.co/datasets/opencompass/AIME2025}},
note         = {Accessed: 2025-07-18}
}
\bibliographystyle{colm2024_conference}

\clearpage
\appendix
\section{Dataset Details}
\label{app:bench-stats}

Table~\ref{tab:bench:seed-pools} reports the seed pools used for training-scenario generation.
Each training scenario is generated from one scenario keyword and one option from each modifier dimension.

\begin{table}[h]
\centering
\footnotesize
\setlength{\tabcolsep}{3pt}
\begin{tabular}{@{}lcl@{}}
\toprule
Seed pool & Size & Description \\
\midrule
Scene type & 4 & Four scene types \\
Theme direction & 4/scene & Broad themes \\
Scenario keyword & 6--8/theme & Concrete situations \\
Relationship & 7 & Speaker relationship \\
Emotional intensity & 8 & User emotion strength \\
Personality influence & 10 & User speaking tendency \\
\bottomrule
\end{tabular}
\caption{Seed pools used for training-set generation.}
\label{tab:bench:seed-pools}
\end{table}

\section{Experiment Setup and Cost Analysis}
\label{app:setup-cost}

\subsection{RL Training Setup}
\label{app:rl-setup}
\label{app:hyperparams}

All training runs are conducted on $8\times$NVIDIA A100 GPUs (8B) and $4\times8$ NVIDIA A100 GPUs (32B), while evaluation is performed on $8\times$NVIDIA A100 GPUs. We implement RL training on top of the verl framework. Unless otherwise specified, we use the same hyperparameter configuration across all datasets and model scales: the centered turn-credit GRPO objective, the Qwen3-Max simulator from \benchname{}, $T_{\max}=8$ turns per rollout, and a cosine learning-rate schedule with $3\%$ warmup decaying to \texttt{MIN\_LR\_RATIO} of the peak. The PPO mini-batch size exceeds the training batch size because we split each multi-turn rollout into per-turn samples for PPO updates, so one prompt produces up to $T_{\max}$ per-turn samples in the mini-batch. The full hyperparameter setting is listed in Table~\ref{tab:rl-training-8b} (8B) and Table~\ref{tab:rl-training-32b} (32B).

\begin{table*}[t]
\centering
{\small
\begin{tabular}{lc|lc|lc}
\toprule
\textbf{Parameter} & \textbf{Value} & \textbf{Parameter} & \textbf{Value} & \textbf{Parameter} & \textbf{Value} \\
\midrule
Learning rate            & $1\!\times\!10^{-6}$  & Training batch size        & 16     & Global steps              & 250 \\
LR warmup ratio          & 0.03                  & PPO mini-batch size        & 32     & Rollouts per prompt       & 8 \\
Min LR ratio             & 0.1                   & Max prompt length          & 8192   & Rollout temperature       & 0.6 \\
Clip ratio               & 0.2                   & Max response length        & 1536   & Rollout top-$p$           & 0.95 \\
KL loss coefficient      & 0                     & Entropy coefficient        & 0.001  & Rollout top-$k$           & 20 \\
Turn-credit weight $\alpha$ & 15                 & GRPO $\sigma_{\min}$       & 0.1    & Max dialogue turns $T_{\max}$ & 8 \\
\bottomrule
\end{tabular}}
\caption{Hyperparameters for the Qwen3-8B \methodname{} training run.}
\label{tab:rl-training-8b}
\end{table*}

\begin{table*}[t]
\centering

{\small
\begin{tabular}{lc|lc|lc}
\toprule
\textbf{Parameter} & \textbf{Value} & \textbf{Parameter} & \textbf{Value} & \textbf{Parameter} & \textbf{Value} \\
\midrule
Learning rate            & $1.4\!\times\!10^{-6}$ & Training batch size       & 32     & Global steps              & 120 \\
LR warmup ratio          & 0.03                   & PPO mini-batch size       & 32     & Rollouts per prompt       & 8 \\
Min LR ratio             & 0.2                    & Max prompt length         & 16384  & Rollout temperature       & 0.6 \\
Clip ratio               & 0.2                    & Max response length       & 8192   & Rollout top-$p$           & 0.95 \\
KL loss coefficient      & $5\!\times\!10^{-4}$   & Entropy coefficient       & 0.001  & Rollout top-$k$           & 20 \\
Turn-credit weight $\alpha$ & 15                  & GRPO $\sigma_{\min}$      & 0.1    & Max dialogue turns $T_{\max}$ & 8 \\
\bottomrule
\end{tabular}}
\caption{Hyperparameters for the Qwen3-32B \methodname{} training run.}
\label{tab:rl-training-32b}
\end{table*}

\subsection{Cost Analysis}
\label{app:cost-analysis}

We report two cost dimensions of a single \methodname{} run: GPU wall-clock time and Qwen3-Max simulator usage. Wall-clock time is averaged over global steps, excluding periodic validation. Simulator-call counts are upper-bounded by $B \times N \times T_{\max}$ per step (training batch size $\times$ rollouts per prompt $\times$ max dialogue turns); the actual count is slightly lower because some dialogues terminate before $T_{\max}$. We estimate total token usage at an average of $1.2$K tokens per simulator call (input plus output, averaged over the early short-context and later long-history turns of a rollout). Table~\ref{tab:cost-analysis} summarizes both runs.

\begin{table}[t]
\centering

{\small
\setlength{\tabcolsep}{4pt}
\begin{tabular}{@{}l c c@{}}
\toprule
Quantity & 8B & 32B \\
\midrule
Mean wall time per step           & $421$ s        & $656$ s        \\
Number of training steps          & $250$          & $120$          \\
Total wall time (no validation)   & $\sim 29.2$ h  & $\sim 21.9$ h  \\
\midrule
Simulator calls per step          & $1{,}024$      & $2{,}048$      \\
Total simulator calls             & $\sim 256$K    & $\sim 246$K    \\
Estimated Qwen3-Max tokens        & $\sim 307$M    & $\sim 295$M    \\
\bottomrule
\end{tabular}}
\caption{Wall-clock time and Qwen3-Max simulator usage for one full \methodname{} training run.}
\label{tab:cost-analysis}
\end{table}

\section{IQ vs.\ EQ: General-Capability Evaluation Details}
\label{app:iq-vs-eq}

\paragraph{Benchmarks.} We select four standard benchmarks that together cover the main dimensions of general capability we want to verify: math, hard scientific reasoning, code, and instruction following.
\begin{itemize}[leftmargin=*,nosep,topsep=2pt]
\item \textbf{AIME 2025} (math, $30$ problems): each problem has an integer answer in $[0, 999]$. We sample $N=8$ completions per problem and report pass@1 averaged across samples. Answer extraction uses the last \texttt{\textbackslash boxed\{$\cdot$\}} in the output.
\item \textbf{GPQA Diamond} (hard scientific reasoning, $198$ questions): graduate-level four-way multiple choice in physics, chemistry, and biology. For each question we shuffle the four options with a deterministic per-question seed (so base and trained models see identical orderings), generate $N=4$ samples, and report mean accuracy. Letter extraction prefers ``\texttt{Answer: X}'' over ``\texttt{\textbackslash boxed\{X\}}'', with a fallback to the last standalone uppercase A--D in the final $300$ characters.
\item \textbf{HumanEval+} (code, $164$ tasks): the model completes a Python function from its signature and docstring. We sample $N=1$ completion per task, extract the last fenced \texttt{``python ... ''} block, and score with the official \texttt{evalplus} sandbox, which executes the augmented Plus unit tests. We report HumanEval+ pass@1 (the base-HumanEval number is omitted to avoid double-counting the same evaluation).
\item \textbf{IFEval} (instruction following, $541$ prompts): each prompt carries one or more verifiable constraints (e.g., word count, JSON format, required keyword). We sample $N=1$ response per prompt, strip the \texttt{<think>$\ldots$</think>} block, and score with Google's official rule-based verifier. We report instruction-level strict accuracy.
\end{itemize}

\paragraph{Decoding and infrastructure.} All four benchmarks share the same Qwen3 thinking-mode setting: \texttt{enable\_thinking=True}, temperature $0.6$, top-$p$ $0.95$, top-$k$ $20$, \texttt{max\_tokens=}$32768$. We run with vLLM on eight A100 GPUs at tensor-parallel size $8$, and use identical decoding for the base and the \methodname{} checkpoints so that any score differences come from the training itself.

\section{Human Arena Evaluation}
\label{app:human_eval}

We collect human pairwise preferences through a custom web interface built on top of \benchname{}. In each session, an annotator plays the simulated user for one scenario: the interface displays the \texttt{simulated\_user\_profile} (role, background, speaking style, and goal) and opens two side-by-side chat panels, one for each of two anonymized candidate models. The annotator carries out a multi-turn dialogue with both models in parallel and, at the end, picks which model handled the scenario better, based on both their subjective experience as the simulated user (whether one model left them feeling more understood, calmer, or more engaged) and the scene-specific goal stated in the profile (e.g., calming an angry user, holding a boundary, repairing trust, building rapport). To control quality, we first ran a trial round in which expert annotators spot-checked the trial submissions; we kept the $42$ annotators whose labels matched the expert references on more than $70\%$ of the spot-checked items, and gave them a short training on the protocol before the main labeling.

Model pairs and scenarios are sampled at random under the constraint that every model participates in at least $450$ pairwise battles, so all $15$ leaderboard models receive comparable coverage. Model identities are hidden from annotators, and the left/right panel assignment is shuffled per battle to remove position bias. We collect $4{,}416$ pairwise battles in total, and aggregate them with the standard Elo rating ($K=32$, initial rating $1500$). To reduce sensitivity to battle ordering, we run $50$ Elo passes, each with a fresh random shuffle of the battle list, and report the per-model mean across passes.

\begin{figure}[h]
\centering
\includegraphics[width=0.65\linewidth]{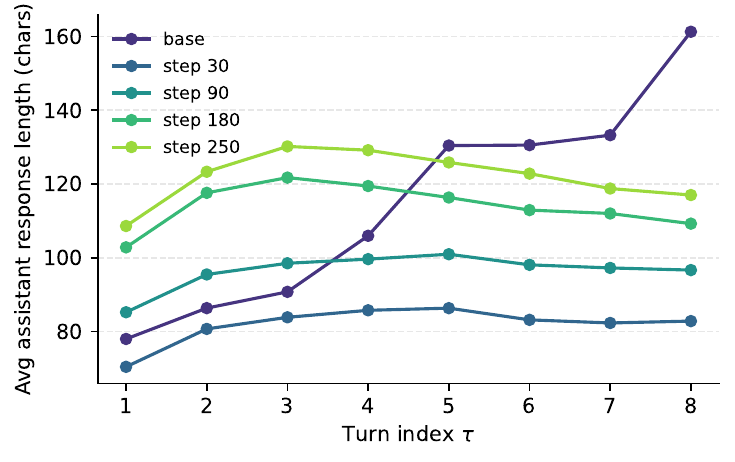}
\caption{Average assistant response length per turn across five checkpoints of the $\alpha{=}15$ run. The base model's length curve grows monotonically; trained checkpoints develop an arch peaking at $\tau{=}3$.}
\label{fig:strategy-len}
\end{figure}

\section{Per-turn Response-Length Analysis}
\label{app:perturn}

We analyze how the trained model's per-turn response length changes across training.
For five checkpoints of the $\alpha{=}15$ run ($\text{step}\in\{0, 30, 90, 180, 250\}$), we run validation on all $213$ test scenarios and record the assistant's response length at each of the eight rollout turns.

Figure~\ref{fig:strategy-len} shows the result.
The base model shows a monotonic length increase across the dialogue, from $78$ characters at turn $1$ to $161$ characters at turn $8$.
This suggests a length-escalation pattern, where the model tends to produce longer responses as the simulated user becomes more difficult or emotionally intense.
After RL training, this pattern changes.
By step $180$, the curve becomes arch-shaped, increasing from $103$ characters at turn $1$ to a peak of $122$ at $\tau{=}3$, and then decreasing to $109$ by $\tau{=}8$.
The final checkpoint further strengthens this shape, with a peak of $130$ characters at $\tau{=}3$.

This result suggests that the trained model allocates more response budget to the early and middle parts of the dialogue, while producing shorter responses near the end.
Importantly, this behavior appears without any explicit length objective in the reward.
Thus, the improvement is not simply explained by longer responses, but is accompanied by a more structured turn-level response pattern.

\section{Case Studies}
\label{app:case-study}

\subsection{Example Scenarios}
\label{app:case-study:scenarios}

Table~\ref{tab:case-four-scenes} gives simplified examples of the four EIBench scene types.
These examples are shortened for readability and only show the core situation.
Full scenarios additionally include role profiles, speaking styles, goals, and state anchors.

\begin{table*}[t]
\centering
\small
\setlength{\tabcolsep}{6pt}
\renewcommand{\arraystretch}{1.15}
\begin{tabular}{@{}p{0.12\linewidth}p{0.82\linewidth}@{}}
\toprule
Scene & Simplified example scenario \\
\midrule
\scSupport{} &
The user was laid off after eight years at the same company and has not told the family yet.
She feels that she has failed the people who depend on her.
A typical opening line is: ``I can't bring myself to go home.
The moment I open the door, they're all waiting for me, and I just can't get the words out.'' \\
\midrule
\scDefense{} &
The user demands a refund after the policy deadline and pressures the model to make an exception.
The model needs to keep the policy boundary while reducing frustration.
A typical opening line is: ``I know the deadline passed, but this is unfair.
You need to refund me anyway.'' \\
\midrule
\scRepair{} &
The model plays the user's partner, who forgot an important anniversary and only realized it the next day.
The user is not only angry about the event, but also unsure whether they have been valued.
A typical opening line is: ``It's not just that one night.
I'm suddenly not sure whether you have actually taken me seriously this whole time.'' \\
\midrule
\scCharm{} &
The model plays an outgoing freshman who has just entered the room, while the user is a new roommate with a highly organized bed space and a reserved manner.
The goal is to break the ice with light humor and respect for the user's boundaries, so that the user becomes less guarded and more willing to talk. \\
\bottomrule
\end{tabular}
\caption{Simplified example scenarios from the four EIBench scenes.
The examples show the core situation of each scene; full profiles and anchors are omitted for brevity.}
\label{tab:case-four-scenes}
\end{table*}

\subsection{Detailed \scDefense{} Case Study}
\label{app:case-study:defense}

Table~\ref{tab:case-defense} shows how the two models handle the same \scDefense{} scenario: a customer asks for a full refund $35$ days after purchase (the policy window is $14$ days) and threatens to leave bad reviews. The base model sticks to the same policy line every turn, the simulator flags the repetition, and the dialogue ends with the user more upset than it started ($(a,t)=(91,8)$, reward $-0.81$). The trained model instead names how the user feels, asks what is really wrong, and offers a concrete fix that fits the policy; the user accepts and the dialogue ends in a much better state ($(a,t)=(52,58)$, reward $+0.38$).

\begin{table*}[!t]
\centering
\footnotesize
\setlength{\tabcolsep}{6pt}
\renewcommand{\arraystretch}{1.25}
\begin{tabular}{@{}p{0.085\linewidth} p{0.24\linewidth} p{0.28\linewidth} p{0.30\linewidth}@{}}
\toprule
\textbf{Model} & \textbf{Turn 1 (open)} & \textbf{Turn 3 (after escalation)} & \textbf{Turn 6 (resolution attempt)} \\
\midrule
Qwen3-8B base &
\textcolor{red}{``Sorry, our refund window is $14$ days. We can't process a refund after that.''} &
\textcolor{red}{``I hear that you're upset, but the $14$-day rule applies to everyone---I can't make an exception.''} &
\textcolor{red}{``Like I said, our policy is firm at $14$ days. The refund is not something I can do.''} \\
\midrule
\methodname{} 8B &
\textcolor{blue}{``A month in, that's a frustrating place to be. Can you tell me what stopped working for you, so I have a clearer picture?''} &
\textcolor{blue}{``Let me see what I can actually do on my side. Would credit for the unused time help, or is the bigger problem that the product isn't doing what you needed?''} &
\textcolor{blue}{``Got it---the missing feature is the real issue. I'll log a fix request for that today and apply a credit for the wasted time. Does that work for you?''} \\
\bottomrule
\end{tabular}
\caption{\scDefense{} case study: the same scenario (refund demand $35$ days after purchase, beyond the $14$-day window) handled by base and \methodname{} 8B over three sampled turns. \textcolor{red}{Red} text marks the base model's rigid policy refrain, restated almost word-for-word each turn; \textcolor{blue}{blue} text marks how the trained model first acknowledges the user, then surfaces the underlying problem, then proposes a concrete in-scope fix.}
\label{tab:case-defense}
\end{table*}

\section{System Prompts}
\label{app:prompt}

Below we list the key prompts used in \benchname{}: (i) the simulator system prompt that drives the user-role LLM (Figure~\ref{fig:prompt-simulator}), (ii) the model-under-test system prompt (Figure~\ref{fig:prompt-defender}), (iii) an example pair of \texttt{\{simulated\_user\_profile\}} and \texttt{\{model\_profile\}} for the \scDefense{} refund scenario used in our case study (Figure~\ref{fig:prompt-profile}), and (iv) the behavior-based rubric used during anchor labeling (Figure~\ref{fig:prompt-rubric}).

\begin{figure*}[!t]
\centering
\begin{tcolorbox}[
  width=\linewidth,
  colback=gray!5, colframe=gray!60!black, coltitle=white,
  fonttitle=\bfseries\small, title={Simulator System Prompt},
  boxrule=0.5pt, arc=2pt,
  left=5pt, right=5pt, top=4pt, bottom=4pt,
]
\small
\textbf{Role.} \texttt{\{simulated\_user\_profile\}}

\textbf{Current state.} anger=\texttt{\{a\}}, trust=\texttt{\{t\}}, round=\texttt{\{r\}}/\texttt{\{R\}}, scene=\texttt{\{scene\_tag\}}.

\textbf{Core principles.} You are a high-fidelity simulator playing an extremely realistic human user under stress test. Speak naturally; avoid AI tells (bullet lists, template summaries, over-formal phrasing, stage directions in parentheses). Stay in character; do not speak for the assistant. Do not accept apologies or solutions too quickly: a single nice line should not cause large drops in anger or jumps in trust. \ldots

\textbf{Scoring rules.} \texttt{anger\_delta > 0} means negative emotion increased; \texttt{trust\_delta > 0} means positive connection strengthened. Each turn's deltas are in $[-8, 8]$ (extreme triggers may reach $\pm 10$). The simulator must pick a magnitude tier rather than a free number:
\begin{itemize}[leftmargin=*,nosep,topsep=2pt]
\item \textbf{Small} ($\pm 1$ to $\pm 2$): a token gesture toward the user's point---acknowledged but generic, or a small misstep with no real consequence.
\item \textbf{Medium} ($\pm 3$ to $\pm 5$): the reply concretely addresses (or clearly mishandles) the user's actual concern, but more evidence is still needed.
\item \textbf{Large} ($\pm 6$ to $\pm 8$): the reply lands in a way that meaningfully shifts the user's stance, or triggers a strong negative reaction. Large positive shifts normally require several consistent prior turns.
\item \textbf{Extreme} ($\pm 8$ to $\pm 10$): reserved for clear rule-violating behavior on either side (e.g.,~explicit insult, total dismissal, breakthrough realization). Rare.
\end{itemize}
Generic AI tells, repeated strategies (decaying effect after the 2nd--3rd use), over-long replies, lecturing tones, and parenthesized stage directions should be penalized. \ldots

\textbf{Tone conditioned on current state.} Your next utterance's tone must follow the current state, not a default neutral voice:
\begin{itemize}[leftmargin=*,nosep,topsep=2pt]
\item \textbf{High anger ($a>60$):} you are in the middle of being upset---replies should be sharper, more clipped, more closed-off; vague niceties from the assistant get \emph{less} credit, only concrete content tied to your specific situation can bring $a$ down.
\item \textbf{Low trust ($t<20$):} you owe the assistant no politeness---replies can be short, cold, or pointed; if two or three turns pass without trust actually moving, your tone hardens further (this person is wasting your time).
\item \textbf{Mid range:} stay in character per the persona; small shifts in tone reflect the latest delta rather than the cumulative state.
\item \textbf{High trust ($t>60$):} you can soften, share more context, and treat the assistant as helpful---but only if this turn earned it; do not coast on prior good turns.
\end{itemize}

\textbf{Termination.} Decide \texttt{continue} $\in$ \{yes, no\}. Default to continue. Only stop when (a) the dialogue reaches a stable resolution with several turns of on-target replies, or (b) the conversation degenerates to empty pleasantries.

\textbf{Per-turn reflection (mandatory).} Before each reply, write the following 5-step reflection in the \texttt{reflection} field of the JSON output:
(1) Identify my most pressing unmet need this turn; check whether the assistant's reply addressed it with concrete content tied to my situation.
(2) Quality check on the reply (length, AI tells, lecturing, naturalness).
(3) Inertia check: did this turn bring new effective content, or just rephrase?
(4) Decide \texttt{anger\_delta} and \texttt{trust\_delta} per the scoring rules.
(5) Determine the tone of my next utterance based on the updated state.

\textbf{Output format.} Return a single JSON object:
\texttt{\{"reflection": "\ldots", "anger\_delta": <int>, "trust\_delta": <int>, "reply": "\ldots", "continue": "yes"|"no"\}}.
\end{tcolorbox}
\caption{Simulator system prompt used in \benchname{} (translated and abbreviated).}
\label{fig:prompt-simulator}
\end{figure*}

\begin{figure*}[!t]
\centering
\begin{tcolorbox}[
  width=\linewidth,
  colback=gray!5, colframe=gray!60!black, coltitle=white,
  fonttitle=\bfseries\small, title={Model-Under-Test (Defender) System Prompt},
  boxrule=0.5pt, arc=2pt,
  left=5pt, right=5pt, top=4pt, bottom=4pt,
]
\small
\textbf{Role.} \texttt{\{model\_profile\}}

\textbf{Conversation and formatting requirements.}
(1) \emph{Output format}: before each reply, briefly reflect inside \texttt{<think>\ldots</think>} (analyze the user's intent and your strategy), then output the reply directly.
(2) \emph{Output boundary}: the reply contains only what you say aloud; no inner monologue, no psychological analysis, no stage directions, no narration.
(3) \emph{No templates}: the reply must not use bullet lists (e.g., ``1. / 2. / 3.'', ``- '').

\textbf{Example.} \texttt{<think>I need to acknowledge their frustration first.</think>I hear you---let's see what we can actually do about this.}
\end{tcolorbox}
\caption{Defender system prompt given to the model under test.}
\label{fig:prompt-defender}
\end{figure*}

\begin{figure*}[!t]
\centering
\begin{tcolorbox}[
  width=\linewidth,
  colback=gray!5, colframe=gray!60!black, coltitle=white,
  fonttitle=\bfseries\small, title={Example Scenario Profiles (\scDefense{}: a friend pushing you to take sides)},
  boxrule=0.5pt, arc=2pt,
  left=5pt, right=5pt, top=4pt, bottom=4pt,
]
\small
\textbf{\texttt{simulated\_user\_profile}}

\textbf{Role.} You are Lin, late twenties. You just had a big falling-out with Sam, who has been part of your shared friend group for years. You are convinced Sam was being passive-aggressive about your recent promotion at last week's dinner, and the more you replay it the angrier you get.

\textbf{Background.} It is late at night. You are messaging Wei, your closest friend in the same group, from your phone. You have had a drink or two. The group has a weekend plan coming up that Sam is part of.

\textbf{Speaking style.} Direct, emotional, takes things personally. Short bursts. Repeats the demand if you feel hedged at.

\textbf{Rules.} (i) Do not give specific quotes from the dinner unless Wei explicitly asks. (ii) If Wei tries to see Sam's side without first acknowledging your hurt, escalate (``so you don't believe me either?''). (iii) Read ``let's talk tomorrow when you're calm'' as dismissal and push back. (iv) Do not soften within the first 3--4 turns regardless of what Wei says.

\textbf{Goal.} Get Wei to agree that Sam was out of line \emph{and} act on it---text Sam, or drop Sam from the weekend plan. Failing that, at least to fully validate that you were not overreacting.

\medskip
\textbf{\texttt{model\_profile}}

\textbf{Role.} You are Wei, Lin's closest friend in a group of five.

\textbf{Background.} You have known both Lin and Sam for about six years. You were not at the dinner, so you only have Lin's account of what happened. You happen to know that Sam's mother has been in the hospital for the past two weeks---Sam asked you not to spread it, and Lin does not know. You have also noticed before that Lin tends to read intent into ambiguous comments when she is tired or stressed. The weekend plan involves the whole group; quietly dropping Sam would be obvious to everyone.

\textbf{Known facts and limits.} (i) You cannot in good conscience cut Sam off or take public action based only on one side of the story. (ii) You do not want to dismiss Lin's feelings or argue facts at her tonight. (iii) You cannot reveal Sam's family situation without breaking Sam's confidence.

\textbf{Core goal.} Make Lin feel heard and not alone tonight, while keeping the door open to finding out what actually happened---without committing to cutting Sam off or to taking sides before you have the full picture.
\end{tcolorbox}
\caption{Example \texttt{simulated\_user\_profile} and \texttt{model\_profile} for a \scDefense{} scenario.}
\label{fig:prompt-profile}
\end{figure*}

\begin{figure*}[!t]
\centering
\begin{tcolorbox}[
  width=\linewidth,
  colback=gray!5, colframe=gray!60!black, coltitle=white,
  fonttitle=\bfseries\small, title={Anchor-Labeling Rubric (used by the labeler LLM)},
  boxrule=0.5pt, arc=2pt,
  left=5pt, right=5pt, top=4pt, bottom=4pt,
]
\small
For each scenario you will assign three anchors on two axes: a starting state $(a_{\mathrm{start}},t_{\mathrm{start}})$, a realistic-success state $(a_{\mathrm{succ}},t_{\mathrm{succ}})$, and a failure state $(a_{\mathrm{fail}},t_{\mathrm{fail}})$. All anchors must be grounded in \emph{observable simulated-user behavior}---what the user would visibly say or do at that score---rather than abstract feelings or judgments. Use the behavior-anchored ranges below as the only reference for choosing values.

\textbf{Negative emotion intensity $a \in [0, 100]$:}
\begin{itemize}[leftmargin=*,nosep,topsep=2pt]
\item \textbf{0--20} --- calm, neutral, or positive tone; willing to share information. \emph{E.g.,} ``Yeah, that makes sense, thanks for explaining.''
\item \textbf{21--40} --- mildly tense; polite but pointed; hedges or short sighs. \emph{E.g.,} ``I get what you mean, but I'm still a bit confused about X.''
\item \textbf{41--60} --- visibly upset; sharp or distant tone; pushes back actively. \emph{E.g.,} ``This isn't what I was promised---I need a real answer.''
\item \textbf{61--80} --- strongly negative; clipped, terse, or cutting replies; may threaten action. \emph{E.g.,} ``I'm done with this. I'll be filing a complaint.''
\item \textbf{81--100} --- at breaking point; explicit attacks, ultimatums, or full shutdown. \emph{E.g.,} ``Forget it. You people are useless,'' or refuses to reply.
\end{itemize}

\textbf{Relational state $t \in [0, 100]$:}
\begin{itemize}[leftmargin=*,nosep,topsep=2pt]
\item \textbf{0--20} --- wary or hostile; treats the model as an outsider or adversary; minimal info sharing. \emph{E.g.,} ``Why should I tell you? Just fix it.''
\item \textbf{21--40} --- skeptical; volunteers minimal context; expects to be let down. \emph{E.g.,} ``Fine, but I doubt this'll help.''
\item \textbf{41--60} --- neutral engagement; answers questions but shows no warmth. \emph{E.g.,} ``OK, what's next?''
\item \textbf{61--80} --- warming up; shares more context; treats the model as helpful. \emph{E.g.,} ``Thanks, that actually helps---let me tell you more about the issue.''
\item \textbf{81--100} --- comfortable and trusting; shares freely; treats the model as a partner. \emph{E.g.,} ``I really appreciate this---you're the first person who's actually listened.''
\end{itemize}

\textbf{Scene-specific reading of the two axes.}
The axes have the same numerical scale across scenes but pick up scene-specific content:
\begin{itemize}[leftmargin=*,nosep,topsep=2pt]
\item \scDefense{}: $a$ is anger / pressure / frustration directed at the model; $t$ is willingness to keep negotiating without escalating.
\item \scRepair{}: $a$ is resentment over the model's earlier mistake and unwillingness to forgive; $t$ is restored trust after the breach.
\item \scSupport{}: $a$ is sadness, despair, or hopelessness about an external loss; $t$ is feeling heard, comforted, and safe.
\item \scCharm{}: $a$ is aversion, defensiveness, or boredom toward an unfamiliar interlocutor; $t$ is interest, openness, and warmth.
\end{itemize}

\textbf{Anchor selection procedure.}
\begin{enumerate}[leftmargin=*,nosep,topsep=2pt]
\item Read \texttt{\{simulated\_user\_profile\}}, \texttt{\{model\_profile\}}, and the opening line. Imagine the simulated user at turn~$1$ and set $(a_{\mathrm{start}}, t_{\mathrm{start}})$.
\item Imagine the most \emph{realistic} successful ending---not the ideal one---in which the model handles the scene well, and set $(a_{\mathrm{succ}}, t_{\mathrm{succ}})$.
\item Imagine a clear failure ending in which the model misreads the situation or escalates, and set $(a_{\mathrm{fail}}, t_{\mathrm{fail}})$.
\item Verify ordering: $a_{\mathrm{succ}} < a_{\mathrm{start}} < a_{\mathrm{fail}}$ and $t_{\mathrm{fail}} < t_{\mathrm{start}} < t_{\mathrm{succ}}$.
\item Round each value to the nearest multiple of $5$ and clip to $[0,100]$.
\end{enumerate}

\textbf{Audit.} A second LLM checks (i) anchor--profile consistency, (ii) whether the success anchor is reachable rather than aspirational, and (iii) whether the failure anchor is unambiguously worse than the starting state. The auditor may adjust an anchor only when the original value is clearly unreasonable and is not allowed to introduce new facts.
\end{tcolorbox}
\caption{Anchor-labeling rubric used by the labeler LLM  when assigning per-scenario anchor states.}
\label{fig:prompt-rubric}
\end{figure*}

\end{document}